\documentclass{acm_proc_article-sp}
\usepackage{algorithm}
\usepackage{algpseudocode}

\usepackage{color}
\usepackage[latin1]{inputenc}
\usepackage{amsmath}
\usepackage{amsfonts}
\usepackage{amssymb}
\usepackage{url}
\usepackage{enumitem}
\usepackage{graphics}
\usepackage{graphicx}
\usepackage{sidecap}
\usepackage{comment}
\usepackage{wrapfig}
\usepackage{amssymb}
\usepackage{amsfonts}
\usepackage{tikz}
 \usepackage{picins}
\usepackage{bigints,multirow,array,booktabs}
\usepackage{anyfontsize}

\usetikzlibrary{arrows,fit,positioning,shapes,snakes}
\let\oldbibliography\thebibliography
\renewcommand{\thebibliography}[1]{%
  \oldbibliography{#1}%
  \setlength{\itemsep}{1pt}%
}

\newdimen\arrowsize
\pgfarrowsdeclare{arcsq}{arcsq}
{
  \arrowsize=0.2pt
  \advance\arrowsize by .5\pgflinewidth
  \pgfarrowsleftextend{-4\arrowsize-.5\pgflinewidth}
  \pgfarrowsrightextend{.5\pgflinewidth}
}
{
  \arrowsize=1.5pt
  \advance\arrowsize by .5\pgflinewidth
  \pgfsetdash{}{0pt} % do not dash
  \pgfsetroundjoin   % fix join
  \pgfsetroundcap    % fix cap
  \pgfpathmoveto{\pgfpoint{0\arrowsize}{0\arrowsize}}
  \pgfpatharc{-90}{-140}{4\arrowsize}
  \pgfusepathqstroke
  \pgfpathmoveto{\pgfpointorigin}
  \pgfpatharc{90}{140}{4\arrowsize}
  \pgfusepathqstroke
}

\usepackage{url}
\usepackage{amsmath}
\usepackage{amssymb}
\usepackage{color}
\usepackage{bigints}
\newcommand{\x}{{\mathbf{x}}}

\newcommand{\ddd}{\mathrm{d}}

\newcommand\independent{\protect\mathpalette{\protect\independenT}{\perp}}
\def\independenT#1#2{\mathrel{\rlap{$#1#2$}\mkern2mu{#1#2}}}

\newtheorem{Defi}{Definition}

\newtheorem{theo1}[Defi]{Theorem}

\begin{document}

\title{Distinguishing Cause from Effect Based on Exogeneity}
\subtitle{[Extended Abstract]
%%\titlenote{A full version of this paper is available as
%%\textit{Author's Guide to Preparing ACM SIG Proceedings Using
%%\LaTeX$2_\epsilon$\ and BibTeX} at
%%\texttt{www.acm.org/eaddress.htm}}
}

\numberofauthors{3} %  in this sample file, there are a *total*
\author{
% 1st. author
\alignauthor
Kun Zhang%\titlenote{Dr.~Trovato insisted his name be first.}
\\
       \affaddr{MPI for Intelligent Systems}\\
       \affaddr{72076 T{\"u}bingen, Germany \&}\\ 
       \affaddr{Info. Sci. Inst., USC}\\
       \affaddr{{\fontsize{9.6}{11.4} \selectfont  4676  Admiralty Way, CA 90292}}\\ 
       \email{kzhang@tuebingen.mpg.de}
% 2nd. author
\alignauthor
Jiji Zhang%\titlenote{The secretary disavows
%any knowledge of this author's actions.}
\\
       \affaddr{Department of Philosophy}\\
       \affaddr{Lingnan University}\\
       \affaddr{Hong Kong S.A.R., China}\\
       \email{jijizhang@ln.edu.hk}
% 3rd. author
\alignauthor Bernhard Sch{\"o}lkopf%\titlenote{This author is the one who did all the really hard work.}
\\
       \affaddr{Max-Planck Institute for Intelligent Systems}\\
       \affaddr{72076 T{\"u}bingen, Germany}\\ 
              \email{bs@tuebingen.mpg.de}
}
% There's nothing stopping you putting the seventh, eighth, etc.
% author on the opening page (as the 'third row') but we ask,
% for aesthetic reasons that you place these 'additional authors'
% in the \additional authors block, viz.
%%\additionalauthors{Additional authors: John Smith (The Th{\o}rv{\"a}ld Group,
%%email: {\texttt{jsmith@affiliation.org}}) and Julius P.~Kumquat
%%(The Kumquat Consortium, email: {\texttt{jpkumquat@consortium.net}}).}
\date{27 Feb. 2015}
% Just remember to make sure that the TOTAL number of authors
% is the number that will appear on the first page PLUS the
% number that will appear in the \additionalauthors section.

\maketitle
\begin{abstract}
Recent developments in structural equation modeling have produced several methods that can usually distinguish cause from effect in the two-variable case. For that purpose, however, one has to impose substantial structural constraints or smoothness assumptions on the functional causal models. % We show that it is possible to perform causal inference directly based on "exogeneity"
In this paper, we consider the problem of determining the causal direction from a related but different point of view, and propose a new framework for causal direction determination.  %We discuss some principles for causal inference, including invariance, exogeneity, and independence between the input distribution and mechanism.
We show that it is possible to perform causal inference based on the condition that the cause is ``exogenous" for the parameters involved in the generating process from the cause to the effect. In this way, we avoid the structural constraints %or smoothness assumptions
required by the SEM-based approaches. %, and provides a fundamental approach for distinguishing cause from effect.
In particular, we
exploit nonparametric methods to estimate marginal and conditional distributions, and propose a bootstrap-based approach to test for the exogeneity condition; the testing results indicate the causal direction between two variables. %, and confirm that causal inference is possible with only the exogeneity condition, without any other constraints on the functional forms.  This is a first work for causal inference that exploits nonparametric models while directly makes use of the ``exogeneity" property of the causal system.
The proposed method is validated on both synthetic and real data.
\end{abstract}

%A category including the fourth, optional field follows...


 % NOT required for Proceedings

\section{Introduction}
Understanding causal relations allows us to predict the effect of changes in a system and control the behavior of the system. Since randomized experiments are usually expensive and often impracticable, causal discovery from non-experimental data has attracted much interest~\cite{Pearl00,Spirtes00}.
To do this, it is crucial to find (statistical) properties in the non-experimental data that give clues about causal relations. For instance, under the causal Markov condition %(NIPS), which states that each variable is independent of all its non-descendants conditional on its parents,
and faithfulness assumption, the causal structure can be partially estimated by constraint-based methods, which make use of conditional independence relationships.% This provides a foundation for constraint-based causal discovery.

Here we are concerned with the two-variable case, in which constraint-based methods, such as the PC algorithm~\cite{Spirtes00}, do not apply. We assume that the given observations are i.i.d., i.e., there is no temporal information. Recently, causal discovery based on structural equation models (SEMs) has proved useful in distinguishing cause from effect~\cite{Shimizu06,Hoyer09,Zhang_UAI09,Zhang09_additive,Mooij10_GPI,Zhangetal15_TIST}; however, the performance of such approaches depends on assumptions on the functional model class and/or on the data-generating functions. On the other hand, there have been attempts in different fields to characterize properties related to causal systems. One such concept (or family of concepts) is known as {\it exogeneity}, which is salient in econometrics~\cite{Engle83_exogeneity,Florens85}. Roughly speaking, the notion expresses the property that the process that determines one variable $X$ is in some sense separate from or independent of the process that determines another variable, say $Y$, given the value of $X$. %In computer science, {\it algorithmic independence}.***** uncomputable...  The above concepts all
 %certain relationships

The sense of ``separateness" or ``independence" in the rough idea has been specified in several ways for different purposes, which result in different concepts of exogeneity. The concept that is most relevant in this paper is the one in the context of model reduction, which was originally proposed as a condition that justifies inferences about the parameters of interest based on the conditional likelihood function rather than the joint likelihood function~\cite{Henry00}. % We will not consider exogeneity as a purely statistical property, but from the causal perspective, as addressed in~\cite{Pearl00,Pearl10_exogeneity}.
Here is the basic idea. Suppose the joint distribution of $(X,Y)$ can be factorized as
\begin{equation} \label{Eq:factorization}
p(X,Y|\theta,\psi) = p(Y|X,\psi)p(X|\theta).
\end{equation}
where the conditional distribution $p(Y|X)$ is parameterized by $\psi$ alone, and the marginal distribution $p(X)$ by $\theta$ alone. According to~\cite{Engle83_exogeneity,Richard80}, $X$ is said to be exogenous for $\psi$ (or any parameter of interest that is a function of $\psi$), if $\psi$ and $\theta$ are variation free\footnote{This is actually the definition of ``weak exogeneity" in~\cite{Engle83_exogeneity}, where three types of exogeneity were defined. Here we consider the i.i.d. case where there is no temporal information, and consequently strong exogeneity in~\cite{Engle83_exogeneity} and weak exogeneity conincide.}, or in other words, are not subject to `cross-restrictions". %; otherwise the range of admissible values for $\psi$ and that for $\theta$ would vary with each other.  %We can interpret exogeneity from either the frequentist or the Bayesian point of view.
From the frequentist point of view, this implies that $\psi$ and $\theta$ are independently estimable: the MLE of $\psi$ and that of $\theta$ are statistically independent according to the sampling distribution. %a weaker condition is that the information matrix, if it exists, is block-diagonal between $\psi$ and $\theta$.
From the Bayesian point of view~\cite{Florens85}, this implies that $\psi$ and $\theta$ are a posteriori independent given independent priors on them.

%\textcolor{red}{[Please check.]} Intuitively, exogeneity describes the invariance property of the conditional probability $P(Y|X)$ to changes in the process generating $X$.  Exogeneity was originally defined in terms of marginal and conditional probabilities (or the involved statistical properties),  whether the exogeneity property holds depends on parametrization.   As pointed out by Pearl~\cite{Pearl00}, to relate exogeneity to causality, one should consider structural parameters (the parameters involved in the structural models) or structural features, instead of statistical ones.  Pearl's definition of exogeneity~\cite[p. 168]{Pearl00} concerns both structural and statistical features, and clearly exhibits causal properties. Suppose $X$ and $Y$ have a direction causal relationship in between without any confounder. The exogeneity condition holds for the true causal direction, and this allows for determining the causal direction based on examining exogeneity.

In this paper we will exploit the above idea to develop a test of whether {\it there exists} a parameterization $(\theta, \psi)$ for $p(X, Y)$ such that $X$ is exogenous for $\psi$, the parameters for $p(Y|X)$. The test is based on bootstrap and is applicable in nonparametric settings. We will also argue that if $X$ is a cause of $Y$ and there is no confounding, then there should exist a parameterization such that $X$ is exogenous for the parameters for $p(Y|X)$. Thus the nonparametric test can be used to indicate the causal direction between two variables, when the test passes for one direction but fails for the other. %We will discuss how the proposed method can be understood from the perspective of structural equations.
Compared to the SEM-based approach, an important novelty of this work is to use exogeneity as a new criterion for causal discovery in general settings, which allows distinguishing cause from effect and detecting confounders without structural constraints on the causal mechanism.\footnote{A related criterion is that of {\it algorithmic independence} between the input distribution $p(X)$ and the conditional $p({Y|X})$ postulated for a causal system $X\rightarrow Y$~\cite{Janzing10}; see also~\cite{Janzing12}.  The algorithmic independence condition is defined in terms of Kolmogorov complexity, which is uncomputable, and the method proposed in this paper provides an alternative way to assess the ``independence" between $p(X)$ and $p(Y|X)$.}

\section{Exogeneity and causality}
%Relating causality to model invariance, exogeneity, information loss, and independence between input distribution and mechanism

% Beyasian counterpart
In this section we define what ``exogeneity" means in this paper, and explain its link to causal asymmetry.
%We begin with definitions of exogeneity and related concepts, and then make their relationships clear.
The concept of exogeneity we will use is adapted from the concept known in econometrics as {\it weak exogeneity}, which is in itself a statistical rather than a causal concept.\footnote{The stronger, causal concept of exogeneity is known as {\it super exogeneity}.} We will show that this statistical notion can nonetheless be exploited to formulate a method that can often determine the causal direction between two variables. %and it was further formulated in the vocabulary of causal reasoning~\cite{Pearl00,Pearl10_exogeneity}.%\footnote{There are two types of analysis of exogeneity~\cite{Florens85}; one considers the inference based on the complete sample results, and the other considers dynamic models where the data were obtained by ``sequential sampling". In this paper we focus on the former scenario.}
%? I consider Pearl's formulation as a further formulation in the vocabulary of causality. Is it fine?

%\textcolor{red}{\it [Maybe we should start with Pearl's exogeneity...  It seems not necessary.]}

\subsection{Exogeneity}\label{Sec:exogeneity}

 The concept of weak exogeneity, as formulated by Engle, Hendry, and Richard (EHR)~\cite{Engle83_exogeneity}, is concerned with when efficient estimation of a set of parameters of interest can be made in a {\it conditional} submodel. For the purpose of this paper, suppose we are given two continuous random variables $X$ and $Y$, on which we have i.i.d. observations that are drawn according to a joint density $p(X, Y|\phi)$. By a reparameterization we mean a one-to-one transformation of the parameter set $\phi$. Our definition below is adapted from the EHR definition, adjusted for our present purpose and setup: %we will simply take $\psi$ as the parameters of interest.  The definition of exogeneity of $X$ for $\psi$ in terms of the notion of classical cut was given in~\cite{Engle83_exogeneity}; see also~\cite{Richard80}.%(NIPS) \footnote{In~\cite{Engle83_exogeneity} three types of exogeneity, namely weak exogeneity, strong exogeneity, and super exogeneity, were defined. Here we consider i.i.d. data, and by default we refer to weak exogeneity. See Footnote 1.}
\begin{Defi}[Exogeneity of $X$ for $p(Y|X)$]  \label{Defi:exo_cut} %% the conditional...
Suppose $p(X,Y)$ is parameterized by $\phi$.  $X$ is said to be exogenous for the conditional $P(Y|X)$ (or simply, exogenous relative to $Y$) if and only if there exists a reparameterization $\phi \rightarrow (\theta, \psi)$, such that
\\{\it (i.)} $p(X,Y|\theta, \psi) = p(Y|X, \psi) p(X|\theta)$, and
\\{\it (ii.)} $\theta$ and $\psi$ are variation free, i.e., $(\theta, \psi) \in \Theta \times \Psi$, where $\Theta$ and $\Psi$ denote the set of admissible values of $\theta$ and $\psi$, respectively.

\begin{comment}
\begin{enumerate}
\item [(i)] $p(X,Y|\theta, \psi) = p(Y|X, \psi) p(X|\theta)$, and
\item [(ii)] $\theta$ and $\psi$ are variation free, i.e., $(\theta, \psi) \in \Theta \times \Psi$, where $\Theta$ and $\Psi$ denote the set of admissible values of $\theta$ and $\psi$, respectively.
\end{enumerate}
\end{comment}
\end{Defi}
Here ``variation free" means that the possible values that one parameter set can take do not depend on the values of the other set. %(NIPS )Under the conditions in Definition~\ref{Defi:exo_cut}, $\theta$ and $\psi$ are said to be orthogonal in~\cite{LANCASTER02}.
Clauses {\it (i.)} and {\it (ii.)} in Definition~\ref{Defi:exo_cut} are the defining conditions for the concept of a {\it (classical) cut}: $[(Y|X;\psi), (X; \theta)]$ is said to operate a (classical) cut on $p(X, Y|\theta, \psi)$ if {\it (i.)} and {\it (ii.)} are satisfied. The cut implies that the maximum likelihood estimates of $\theta$ and $\psi$ can be computed from $p(X|\theta)$ and $p(Y|X, \psi)$, respectively, and so the MLEs $\hat{\theta}$ and $\hat{\psi}$ are independent according to the sampling distribution. %(NIPS) (A weaker condition is that the information matrix w.r.t. $(\theta, \psi)$ is block-diagonal.)
The concept of exogeneity formalizes the idea that the mechanism generating the exogenous variable $X$ does not contain any relevant information about the parameter set $\psi$ for the conditional model $p(Y|X)$. %Therefore, under the exogeneity condition of $X$ for $\psi$, when estimating $\psi$, we can treat $X$ as if it was not random. %(NIPS), i.e., we do not care about how $X$ was generated or what $p_X$ is; the estimator obtained by maximizing the conditional likelihood does not suffer from any loss of efficiency compared to maximizing the complete likelihood.

The concept of cut also has a Bayesian version:~\cite{Florens85,Mouchart04}. %(NIPS) and allows an admissible reduction of the complete model $p(X,Y|\theta, \psi)$ to the conditional model $p(Y|X, \psi)$, in that both models lead to the same posterior distribution of the parameter set $\psi$~\cite{Florens85,Mouchart04}.
\begin{comment}
Below we give the definition of mutual exogneneity according to~\cite{Florens85}.

\begin{Defi}[Mutual exogeneity] \label{Def:mutual_exo}
$X$ and $\psi$ are mutually exogenous if and only if
\begin{enumerate}
\item [(i)] $\psi$ and $X$ are independent, i.e., $\psi \independent X$, and
\item [(ii)] $\psi$ is sufficient in the conditional distribution of $Y$ given $X$, i.e., $\theta \independent Y|(\psi,X)$.
\end{enumerate}
\end{Defi}
Here condition {\it (i)} is to do with the independence between $\psi$ and $X$; those two quantities play different roles in the model $p(X,Y,\psi, \theta)$, and consequently this independence condition is usually not convenient to verify.  Moreover, for the same reason, there is no fully equivalent concept in sampling theory (it is weaker than exogeneity defined in Definition~\ref{Defi:exo_cut}, because the property of $\theta$ is not specified).
\end{comment}
%A natural way of obtaining the mutual exogeneity of $X$ and $\psi$ is to exploit a stronger but more operational condition, namely the condition of the Bayesian cut.
%This can be verified with the condition of Bayesian cut~\cite{Florens90}.
% Definition of Bayesian cut []...
\begin{Defi}[Bayesian cut]\label{Def:cut}
$[(Y|X;\psi), (X; \theta)]$ operates a Bayesian cut on $p(X,Y|\theta,\psi)$ if
\\{\it(i.)} $\psi$ and $\theta$ are independent {\it a priori}, i.e., $\psi \independent \theta$,
\\{\it (ii.)} $\theta$ is sufficient for the marginal process of generating $X$, i.e., $\psi\independent X|\theta$, and
\\{\it(iii.)} $\psi$ is sufficient for the conditional process of generating $Y$ given $X$, i.e., $\theta \independent Y|(\psi,X)$.

\begin{comment}
\begin{enumerate}
\item [(i)] $\psi$ and $\theta$ are independent {\it a priori}, i.e., $\psi \independent \theta$,
\item [(ii)] $\theta$ is sufficient in the sampling marginal process of $X$, i.e., $\psi\independent X|\theta$, and
\item [(iii)] $\psi$ is sufficient in the conditioal distribution of $Y$ given $X$, i.e., $\theta \independent Y|(\psi,X)$.
\end{enumerate}
\end{comment}
\end{Defi}

A Bayesian cut allows a complete separation of inference (on $\theta$) in the marginal model and of inference (on $\psi$) in the conditional model. The prior independence between $\theta$ and $\psi$ in the Bayesian cut is a counterpart to the variation-free condition in the classical cut, and the last two conditions in Definition~\ref{Def:cut} implies condition {\it (i.)} in Definition~\ref{Defi:exo_cut}. Thus, the Bayesian cut is equivalent to the classical cut in sampling theory, and for the purpose of this paper can be regarded as interchangeable. Therefore, the exogeneity of $X$ relative to $Y$ can also be defined as that there exists a reparameterization $(\theta, \psi)$ of $p(X,Y)$ such that $[(Y|X;\psi), (X; \theta)]$ operates a Bayesian cut on $p(X,Y|\theta,\psi)$.  %We note that
%Note that in Economics sometimes the following condition in the above definition is replaced by tha condition that $\psi$ and $\theta$ are variation free.

\begin{comment}
The following theorem, extracted from~\cite{Florens90}, relates the Bayesian cut to the independence of the parameters according to the posterior distribution, as well as mutual exogeneity.
% also refer to book "Bayesian Inference in Dynamic Econometric Models"
\begin{theo1} \label{Theo:prior_posterior}
If $[\psi, (X, \theta)]$ operates a Bayesian cut in $p(X,Y,\{\psi,\theta\})$, then
\begin{itemize}
\item [(i)] $X$ and $\psi$ are mutually exogenous, and
\item [(ii)] $\psi$ and $\theta$ are independent a posteriori.
\end{itemize}
On the other hand, if $X$ and $\psi$ are mutually exogenous and if $\theta \independent \psi |X$, $[\psi, (X, \theta)]$ operates a Bayesian cut.
\end{theo1}
\end{comment}

\subsection{Possible Situations Where the Parameterization Fails to Operate a Bayesian Cut}
\begin{figure}[htp]
\begin{center}
\begin{tikzpicture}[scale=1, line width=0.5pt, inner sep=0.2mm, shorten >=1pt, shorten <=1pt]
  \scriptsize
\draw (0,0) node(1) [circle, draw] {{\;$\theta$\;}};
  \draw (1.0, 0) node(2) [circle, fill=black!25, draw] {{\;$x_i$\;}};
\draw (2.0, 0) node(3) [circle, fill=black!25, draw] {{\;$y_i$\;}};
  \draw (1.4,1) node(4) [circle, draw] {{\;$\psi$\;}};
  \draw[-arcsq] (1) -- (2); %node[pos=.5,above] {{\large$a_1$}};
  \draw[-arcsq] (2) -- (3); %node[pos=.5,above] {{\large$a_2$}};
  \draw[-arcsq] (4) -- (3);
\node[rectangle, inner sep=0.4mm, fit= (2) (3),label=below right:{$n$}, xshift=5.4mm] {};
\node[rectangle, inner sep=3.2mm, draw=black!100,fit= (2) (3)] {};
\end{tikzpicture}
~~~~\begin{tikzpicture}[scale=1, line width=0.5pt, inner sep=0.2mm, shorten >=1pt, shorten <=1pt]
  \scriptsize
\draw (0,0) node(1) [circle, draw] {{\;$\theta$\;}};
  \draw (1.0, 0) node(2) [circle, fill=black!25, draw] {{\;$x_i$\;}};
\draw (2.0, 0) node(3) [circle, fill=black!25, draw] {{\;$y_i$\;}};
  \draw (1.4,1.0) node(4) [circle, draw] {{\;$\psi$\;}};
\draw (0.3,0.9) node(5) [circle, draw] {{\;$\gamma$\;}};
  \draw[-arcsq] (1) -- (2); %node[pos=.5,above] {{\large$a_1$}};
  \draw[-arcsq] (2) -- (3); %node[pos=.5,above] {{\large$a_2$}};
  \draw[-arcsq] (4) -- (3);
\draw[-arcsq] (5) -- (1);
\draw[-arcsq] (5) -- (4);
\node[rectangle, inner sep=0.4mm, fit= (2) (3),label=below right:{$n$}, xshift=5.4mm] {};
\node[rectangle, inner sep=3.2mm, draw=black!100,fit= (2) (3)] {};
\end{tikzpicture}
 \\(a)~~~~~~~~~~~~~~~~~~~~~~~~~~~(b)\\~~\\
\begin{tikzpicture}[scale=1, line width=0.5pt, inner sep=0.2mm, shorten >=1pt, shorten <=1pt]
  \scriptsize
\draw (0,0) node(1) [circle, draw] {{\;$\theta$\;}};
  \draw (1.0, 0) node(2) [circle, fill=black!25, draw] {{\;$x_i$\;}};
\draw (2.0, 0) node(3) [circle, fill=black!25, draw] {{\;$y_i$\;}};
  \draw (1.4,1.0) node(4) [circle, draw] {{\;$\psi$\;}};
\draw (0.3,0.9) node(5) [circle, draw] {{\;$\gamma$\;}};
  \draw[-arcsq] (1) -- (2); %node[pos=.5,above] {{\large$a_1$}};
  \draw[-arcsq] (2) -- (3); %node[pos=.5,above] {{\large$a_2$}};
  \draw[-arcsq] (4) -- (3);
\draw[-arcsq] (5) -- (4);
\draw[-arcsq] (5) -- (2);
\node[rectangle, inner sep=0.4mm, fit= (2) (3),label=below right:{$n$}, xshift=5.4mm] {};
\node[rectangle, inner sep=3.2mm, draw=black!100,fit= (2) (3)] {};
\end{tikzpicture}
~~~~\begin{tikzpicture}[scale=1, line width=0.5pt, inner sep=0.2mm, shorten >=1pt, shorten <=1pt]
 \scriptsize
\draw (0,0) node(1) [circle, draw] {{\;$\theta$\;}};
  \draw (1.0, 0) node(2) [circle, fill=black!25, draw] {{\;$x_i$\;}};
\draw (2.0, 0) node(3) [circle, fill=black!25, draw] {{\;$y_i$\;}};
  \draw (1.4,1.0) node(4) [circle, draw] {{\;$\psi$\;}};
\draw (0.3,0.9) node(5) [circle, draw] {{\;$\gamma$\;}};
  \draw[-arcsq] (1) -- (2); %node[pos=.5,above] {{\large$a_1$}};
  \draw[-arcsq] (2) -- (3); %node[pos=.5,above] {{\large$a_2$}};
  \draw[-arcsq] (4) -- (3);
\draw[-arcsq] (5) -- (1);
\draw[-arcsq] (5) -- (3);
\node[rectangle, inner sep=0.4mm, fit= (2) (3),label=below right:{$n$}, xshift=5.4mm] {};
\node[rectangle, inner sep=3.2mm, draw=black!100,fit= (2) (3)] {};
\end{tikzpicture}
\\(c)~~~~~~~~~~~~~~~~~~~~~~~~~~~(d)
\end{center}
\caption{Graphical representation of the data-generating process. (a)  $[(Y|X;\psi), (X; \theta)]$ operates a Bayesian cut (implying that $X$ and $\psi$ are mutually exogenous). (b), (c), and (d) correspond to three situations where $[(Y|X;\psi), (X; \theta)]$ does not operate a Bayesian cut: (b) $\psi$ and $\theta$ are dependent {\it a priori}, as both of them depend on $\gamma$, which is a function of $\theta$ or $\psi$; (c) $\theta$ is not sufficient in modeling the marginal distribution of $X$, where $\gamma$ is a function of $\psi$; (d) $\psi$ is not sufficient in modeling the conditional distribution of $Y$ given $X$, where $\gamma$ is a function of $\theta$.}
\label{fig:data_generating}\end{figure}
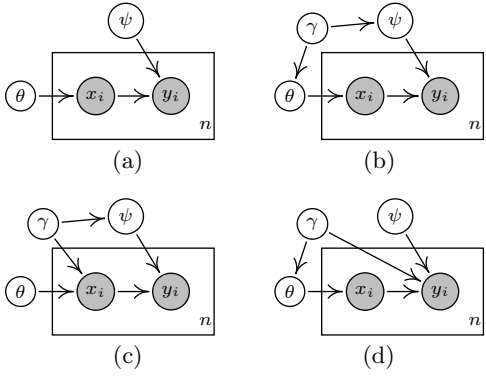

Fig.~\ref{fig:data_generating}(a) shows a data-generating process of $X$ and $Y$ from where $[(Y|X;\psi), (X; \theta)]$ operates a Bayesian cut. Note that in Definition~\ref{Def:cut}, the two requirements of sufficiency of $\psi$ and $\theta$ for the marginal and the conditional (conditions {\it(ii.)} and {\it(iii.)}), respectively, are only restrictive under the assumption of prior independence of $\theta$ and $\psi$ (condition {\it(i.)}); otherwise, conditions {\it(ii.)} and {\it(iii.)} can be trivially met by, for example, taking $\theta$ and $\psi$ to be the same. In fact, any two conditions in Definition~\ref{Def:cut} could be trivial, given that the other does not hold.
%When one (or more) condition in Definition~\ref{Def:cut} is violated, $[\psi, (X,\theta)]$ does not operate a Baysian cut, i.e., $X$ is not exogenous for $\psi$.
Fig.~\ref{fig:data_generating}(b--d) shows the situations where conditions {\it(i.)}, {\it(ii.)}, and {\it(iii.)} are violated, respectively. %(NIPS), so that $[\psi, (X,\theta)]$ does not operate a Baysian cut.
%Note that by reparameterization, the three situations can reduce to one another. %(NIPS) Take situations (b) and (c) as an example. If we divide $\theta$ in (b) into $(\theta_\gamma, \theta_{\independent})$, where $\theta_\gamma$ depends on $\gamma$ while $\theta_{\independent}$ does not, and consider $\theta_{\independent}$ as the new $\theta$, (b) becomes (c). Similarly, if we merge $\gamma$ and $\theta$ in (c) as the new $\theta$, we then have (b).
In all those situations, $\theta$ and $\psi$ are not independent {\it a posteriori}. %or the maximum likelihood estimators $\hat{\theta}$ and $\hat{\psi}$ are not independent according to the sampling distribution.

\begin{comment}
Definition of exogeneity: Pearl's definition, Engle's definition...

In particular, suppose we are interested in the parameters $\psi$ in the conditional distribution $p(Y|X, \psi)$. Let $Z \triangleq (X,Y)$. We have
$$p(X,Y|\{\theta, \psi\}) = p(Z|\{\theta, \psi\}) = p(Z|X, \psi) \cdot p(X| \theta) = p(Y|X, \psi) \cdot p(X| \theta).$$
One can then consider the parameters $\theta$ in $p(X; \theta)$ as {\it nuisance parameters}. Information loss...

There are some closely related concepts in statistical inference.

The definition of ``variation free" according to the space []...

To make it more precise,~\cite{Engle83_exogeneity} defined exogeneity in terms of the concept ``cut".  Definition of Bayesian cut []...

Admissibility []...

(local or general) orthogonality of parameters...
\end{comment}

\subsection{Relation to Causality} \label{sec:example}

As Pearl~\cite{Pearl00} rightly stressed, the EHR concept of weak exogeneity is a statistical rather than a causal notion. Unlike the concept of super exogeneity, it is not defined in terms of interventions or multiple regimes. That is why, as we will show, the hypothesis that $X$ is exogenous relative to $Y$ in the sense we defined is generally testable by observational data. However, it is also linked to causality in that it is \textcolor {black}{arguably} a necessary condition for an unconfounded causal relation: if $X$ is a cause of $Y$ and there is no common cause of $X$ and $Y$, then $X$ is exogenous relative to $Y$ in the sense we defined.\footnote{In this paper we use "unconfounded" to mean the absence of any common cause.} This follows from the principle we indicated at the beginning: if $X$ is an unconfounded cause of $Y$, then the process or mechanism that determines $X$ is separate or independent from the process or mechanism that determines $Y$ given $X$. The separation of processes ensures the {\it existence} of separate parameterizations of the processes, which will then satisfy our definition of exogeneity.    

\textcolor {black}{We have argued that if $X$ and $Y$ are causally related and unconfounded, the exogeneity property holds for the correct causal direction.  Furthermore, if it turns out that there is one and only one direction that admits exogeneity, then the direction for which the exogeneity property holds must be the correct causal direction.} This suggests the following approach to inferring the causal direction between $X$ and $Y$ based on some tests of exogeneity, assuming \textcolor {black}{that $X$ and $Y$ are causally related and} that there is no common cause of $X$ and $Y$ (or in other words, $X$ and $Y$ form a {\it causally sufficient} system): test whether (1) $X$ is exogenous for $p(Y|X)$ and whether (2) $Y$ is exogenous for $p(X|Y)$, and if one of them holds and the other does not, we can infer the causal direction accordingly. Of course it may also turn out that neither (1) nor (2) holds, which will indicate that the assumption of causal sufficiency is not appropriate, or that both (1) and (2) hold, which will indicate that the causal direction in question is not identifiable by our criterion.\footnote{Note that we are not concerned with the case in which $X$ and $Y$ are not causally connected and hence statistically independent; in that case, exogeneity trivially holds in both directions.} 

%%%%%

% As Pearl~\cite{Pearl00} rightly stressed, the EHR concept of weak exogeneity is a statistical rather than a causal notion. Unlike the concept of super exogeneity, it is not defined in terms of interventions or multiple regimes. That is why, as we will show, the hypothesis that $X$ is exogenous relative to $Y$ in the sense we defined is generally testable by observational data. However, it is also linked to causality because it is a necessary condition for an unconfounded causal relation: if $X$ is an unconfounded cause of $Y$ (i.e., if $X$ is a cause of $Y$ and there is no common cause of $X$ and $Y$), then $X$ is exogenous relative to $Y$ in the sense we defined. \textcolor{red}{If we assume that $X$ and $Y$ are causally related without a confounder,} this suggests the following approach to inferring the causal direction between $X$ and $Y$ based on some tests of exogeneity: test whether (1) $X$ is exogenous for $p(Y|X)$ and whether (2) $Y$ is exogenous for $p(X|Y)$, and if one of them holds and the other does not, we can infer the causal direction accordingly. Of course it may also turn out that neither (1) nor (2) holds, which will indicate the presence of confounding, or that both (1) and (2) hold, in which case the causal direction is not identifiable based on the criterion of exogeneity.

A familiar example of a non-identifiable situation is when $X$ and $Y$ follow a bivariate normal distribution. In that case, as shown by EHR~\cite{Engle83_exogeneity}, there is a cut $[(Y|X;\psi), (X; \theta)]$ in one direction, as well as a cut $[(X|Y;\tilde{\psi}), (Y; \tilde{\theta})]$ in the other. Below we give an example where the causal direction is identifiable based on exogeneity.

{\bf An example of identifiable situation: Linear non-Gaussian case.} 

Let $X$ follow a Gaussian mixture model with two Gaussians, $X \sim \sum_{i=1}^2\pi_i\mathcal{N}(u_i, \sigma_i^2)$, where $\pi_i >0$ and $\pi_1+\pi_2 = 1$, and let $Y = c + \beta X + E$ where $E \sim \mathcal{N}(0,\sigma^2)$. Therefore $\theta = \{\pi_i, \mu_i, \sigma_i\}_{i=1}^2$ and $\psi = \{c, \beta, \sigma^2  \}$. We then have
\begin{flalign} \nonumber
p(X,Y|\theta, \beta) &= \sum_i\pi_i \mathcal{N}(x;\mu_i,\sigma_i^2)\mathcal{N}(y; c + \beta x,\sigma^2) \\ \nonumber &
= \sum_i\pi_i \mathcal{N}(y;\tilde{\mu}_i, \tilde{\sigma}_i^2)\mathcal{N}(x;\tilde{c}_i + \tilde{\beta}_i y, \gamma_i^2),
\end{flalign}
where $\tilde{\mu}_i = c + \beta \mu_i$, $\tilde{\sigma}_i^2 = \beta^2 \sigma_i^2 + \sigma^2$, $\tilde{c}_i = \frac{\mu_i \sigma^2 - c\beta \sigma_i^2}{\beta \sigma_i^2 + \sigma^2}$, $\tilde{\beta}_i = \frac{\beta \sigma_i^2}{\beta \sigma_i^2 + \sigma^2}$, and $\gamma_i^2 = \frac{\sigma^2 \sigma_i^2}{\beta \sigma_i^2 + \sigma^2 }$.
That is,
\begin{flalign} \nonumber
 Y &\sim \sum_{i=1}^2 \pi_i\mathcal{N}(y;\tilde{u}_i, \tilde{\sigma}_i^2), \textrm{~~and~~}  \\ \nonumber
p(X|Y,\theta,\psi) &
= \sum_{i} \frac{\pi_i \mathcal{N}(y;\tilde{\mu}_i,\tilde{\sigma}_i^2)}{p(Y|\theta, \psi)} \cdot \mathcal{N}(x;\tilde{c}_i + \tilde{\beta}_i y, \gamma_i^2).
\end{flalign}
%\piccaption{An illutration on the identifiability of a linear non-Gaussian model based on ``exogeneity". $X$ is generated by a mixture of two Gaussians, and $Y$ is generated by $Y=X+E$, where $E\sim \mathcal{N}(0,1)$. Here $X$ is exogenous for parameters in $p_{Y|X}$, while $Y$ is not exogenous for parameters in $p_{X|Y}$.\label{fig:illust_asymmetry}}
%\parpic(2.5in,2.0in)[r]{%
%\includegraphics[width = .45\linewidth]{fig/illustrate_exogeneity.pdf}}

\begin{figure}
\begin{center}
\includegraphics[width = .85\linewidth]{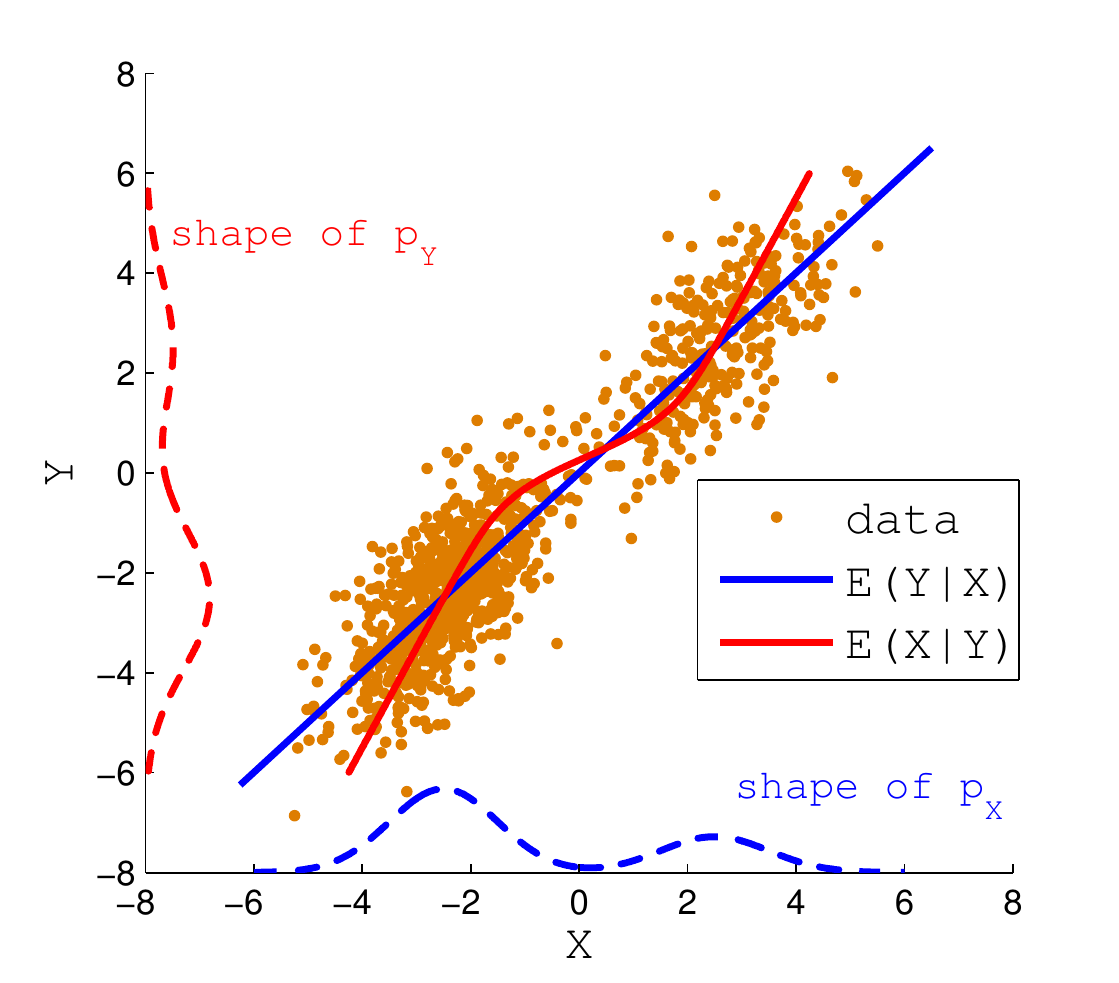}
\caption{\protect An illutration on the identifiability of a linear non-Gaussian model based on ``exogeneity". $X$ is generated by a mixture of two Gaussians, and $Y$ is generated by $Y=X+E$, where $E\sim \mathcal{N}(0,1)$. Here $X$ is exogenous for parameters in $p_{Y|X}$, while $Y$ is not exogenous for parameters in $p_{X|Y}$.\label{fig:illust_asymmetry}}
\end{center}
\end{figure}

Clearly, if $\pi_1\pi_2 \neq 0$, no matter how one parametrizes the density of $Y$, the conditional distribution of $X$ given $Y$ would involves those parameters that model the marginal density of $Y$. The sufficient parameter set of the distribution of $Y$, $\tilde{\theta}$, and that of the conditional distribution of $X$ given $Y$, $\tilde{\psi}$, cannot be variation-free or independent {\it a priori}; see Fig.~\ref{fig:data_generating}(b).  Alternatively, one can keep those parameters that are independent {\it a priori} from $\tilde{\theta}$ in $\tilde{\psi}$, i.e., $\tilde{\psi}$ and $\tilde{\theta}$ become independent {\it a priori}, but $\tilde{\psi}$ is then not sufficient in modeling $p(X|Y)$; see Fig.~\ref{fig:data_generating}(d). In both situations $Y$ is not exogenous for $\tilde{\psi}$. Hence in this linear non-Gaussian case the exogeneity condition only holds for the direction $X\rightarrow Y$, and the causal direction is identifiable. Fig.~\ref{fig:illust_asymmetry} gives an intuitive illustration on how the shape of $P(Y)$ and that of $E(X|Y)$, which is determined by $P(X|Y)$, are related.

\section{Causal direction determination by testing for exogeneity with bootstrap} \label{Sec:bootstrap}

We now describe our approach to testing exogeneity. We will first illustrate how bootstrap can be used to test whether a given parametric model constitutes a (Bayesian) cut, and then develop a nonparametric test for exogeneity based on bootstrap.

 %Suppose that the marginal and conditional models $p(X|\theta)$ and $p(Y|X, \psi)$ are given.  If we can draw random samples from the population of $(X,Y)$, we can easily verify if the $\theta$ and $\psi$ are dependent {\it a posteriori} (suppose both of them have flat priors). However, it is not possible to directly access the population; fortunately, when the observations can be assumed to be drawn from an i.i.d. population, this can be approximately achieved by bootstrap~\cite{Efron93}.
%The basic idea behind bootstrap is to use the empirical data distribution of the sample to replace (or mimic) the population distribution, from which one can draw random samples. %Roughly speaking, there are two different approaches to bootstrap resampling---sampling the residuals produced by a given regression model and sampling the data pairs $(x_i,y_i)$. They apply to different problems. For instance, when using bootstrap to assess the properties of an estimator in the regression model, where $X$ is the predictor and $Y$ is the response, we can sample the regression residuals because the values of $X$ are fixed. However, in our case, both $X$ and $Y$ are random variables, and we are interested in the process that generates both of them. Therefore, we have to
%In particular, we use paired bootstrap, that is, we draw the data point from the given pairs $(x_i,y_i)$ by random sampling with replacement.

\subsection{Bootstrap-Based Test for Bayesian Cut in the Parametric Case}
%{\bf Starting with parametric models. }
In this section, we assume that a parametric form $p(X,Y|\theta,\psi) = p(X|\theta)p(Y|X,\psi)$ is given.
%$$p_{XY}(x,y;\theta, \psi) = p_X(x; \theta) p_{Y|X}(y|x; \psi).$$
We would like to see whether the estimates of $\theta$ and of $\psi$ in (\ref{Eq:factorization}) are independent, according to the sampling distribution; in other words, with a noninformative prior, we want to test if the posterior distribution $p(\theta, \psi|\mathcal{D})$ has no coupling between $\theta$ and $\psi$. In this case we are examining if $[(Y|X;\psi), (X;\theta)]$ operates a Bayesian cut.

%\textcolor{red}{[Such a approach has been used in the literature to assess the dependence, as well as uncertainty, in the parameter estimates according to the sampling distribution; for instance, see~\cite[Sec. 5.7]{Efron82}]}
\begin{small}
\begin{table*}[htp]\vspace{-.3cm}
\centering
\caption{Notation involved in the proposed method based on exogeneity and bootstrap}
% \resizebox{1\textwidth}{!}{\begin{minipage}{\textwidth}
\begin{tabular}{l|p{13.0cm} }
\toprule
$(\mathbf{x},\mathbf{y})$ & given sample of $(X,Y)$ \\
$(\mathbf{x}^{*(b)}, \mathbf{y}^{*(b)})$ & $b$th bootstrap resample \\
$\hat{\theta}^{*(b)}, \hat{\psi}^{*(b)}$ & estimate of parameters $\theta$ and $\psi$ on $(\mathbf{x}^{*(b)}, \mathbf{y}^{*(b)})$ \\
$\hat{p}^{*(b)}(X=\tilde{\mathbf{x}})$ & marginal densities estimated on $(\mathbf{x}^{*(b)}, \mathbf{y}^{*(b)})$ evaluated at $X=\tilde{\mathbf{x}}$ \\
$\hat{p}^{*(b)}(Y|X=\tilde{\mathbf{x}})$ & conditional densities estimated on $(\mathbf{x}^{*(b)}, \mathbf{y}^{*(b)})$ evaluated at $X=\tilde{\mathbf{x}}$ \\
$H_{Y|X}^{*(b)}(\tilde{\mathbf{x}})$ & quantity associated with $\hat{p}^{*(b)}(Y|X=\tilde{\mathbf{x}})$, defined as $\mathbb{E}_{Y|X} \log \hat{p}^{*(b)}(Y|X=\tilde{\mathbf{x}})$ on $(\mathbf{x}^{*(b)}, \mathbf{y}^{*(b)})$\\
\bottomrule
\end{tabular}
\label{tab:notation}
% \end{minipage} }
\end{table*}
\end{small}

Bootstrap has been used in the literature to assess the dependence, as well as uncertainty, in the parameter estimates according to the sampling distribution; see e.g.~\cite[Sec. 5.7]{Efron82}. For clarity, Table~\ref{tab:notation} gives the notation used in the proposed bootstrap-based method.
Suppose we draw bootstrap resamples $(\mathbf{x}^{*(b)}, \mathbf{y}^{*(b)})$, $b=1,...,B$, from the original sample $(\mathbf{x}, \mathbf{y} ) = (x_i, y_i)_{i=1}^N$ with paired bootstrap, i.e.,
each resample $(\mathbf{x}^{*(b)}, \mathbf{y}^{*(b)})$ is obtained by independently drawing $N$ pairs from the original sample with replacement.
 On each of them, we can calculate the parameter estimates $\hat{\theta}^{*(b)}$ and $\hat{\psi}^{*(b)}$. The independence between $\theta$ and $\psi$ according to the sampling distribution is then transformed to statistical independence between the bootstrap estimates $\hat{\theta}^{*(b)}$ and $\hat{\psi}^{*(b)}$, $b=1,...,B$. To assess the latter, any independence test method, such as the correlation test, would apply.

\subsection{Bootstrap-Based Test for Exogeneity in the Nonparametric Case} \label{Sec:Para_to_Nonpara}
%{\bf With nonparametric models for $p({Y|X})$. }

Let $\tilde{\mathbf{x}}$ be a fixed set of values of $X$, and $\tilde{x}_i$ be a point in $\tilde{\mathbf{x}}$. $\tilde{\mathbf{x}}$ can be drawn from the given data set, or randomly sampled on the support of $X$, given that it contains enough points such that the values of $P(X)$ and $p(Y|X)$ evaluated at $\tilde{\mathbf{x}}$ well approximate the continuous densities.  In our experiments we used 80 evenly-spaced sample points between the minimum and maximum values of $X$ as $\tilde{\mathbf{x}}$ (so its length is $N=80$).

On the bootstrap resamples, $\log \hat{p}^{*(b)}({X}=\tilde{\mathbf{x}})$ is fully determined by $\hat{\theta}^{*(b)}$; similarly, $\log \hat{p}^{*(b)}(Y|X=\tilde{\mathbf{x}})$ is a function of $\hat{\psi}^{*(b)}$, and so is the quantity $H_{Y|X}^{*(b)}(\tilde{\mathbf{x}}) \triangleq \mathbb{E}_{Y|X} \log \hat{p}^{*(b)}(Y|X=\tilde{\mathbf{x}})$. Note that $\hat{p}^{*(b)}(Y|X=\tilde{x}_i)$ is the estimated distribution of $Y$ at $X=\tilde{x}_i$, and hence $H_{Y|X}^{*(b)}(\tilde{\mathbf{x}})$ can be considered as negative entropies of $Y$ on the $b$th bootstrap resample evaluated at $X=\tilde{\mathbf{x}}$.

Suppose all involved parameters are identifiable, i.e., the mappings $\theta \mapsto p(X|\theta)$ and $\psi \mapsto p(Y|X, \psi)$ are both one-to-one~\cite{Lehmann98}. Then the mapping between $\hat{\theta}^{*(b)}$ and $\log \hat{p}^{*(b)}({X}=\tilde{\mathbf{x}})$ and that between $\hat{\psi}^{*(b)}$ and $\log \hat{p}^{*(b)}(Y|X=\tilde{\mathbf{x}})$ are both one-to-one. Hence, the independence between $\hat{\theta}^{*(b)}$ and $\hat{\psi}^{*(b)}$, $b=1,...,B,$ implies that between $\log \hat{p}^{*(b)}({X}=\tilde{\mathbf{x}})$ and $H_{Y|X}^{*(b)}(\tilde{\mathbf{x}})$.

As a consequence, in nonparametric settings, we can imagine that there exist effective parameters $\theta$ and $\psi$, and can still assess where they follow a Bayesian cut by testing for independence between the bootstrapped estimates $\log \hat{p}^{*(b)}({X}=\tilde{\mathbf{x}})$ and $H_{Y|X}^{*(b)}(\tilde{\mathbf{x}})$.
% the exogeneity condition of $X$ for ``parameters" in $P(Y|X)$ by examining the independence between $\log \hat{p}^{*(b)}(X=\tilde{\mathbf{x}})$ and $H_{Y|X}^{*(b)}(\tilde{\mathbf{x}})$.
Note that in the nonparametric case, the ``parameters" $\theta$ and $\psi$ are not observable. The previous argument shows that if there exists $(\theta,\psi)$ admitting a Bayesian cut, $\log \hat{p}^{*(b)}({X}=\tilde{\mathbf{x}})$ and $H_{Y|X}^{*(b)}(\tilde{\mathbf{x}})$ are independent; otherwise they are always dependent. In words, testing for independence between the bootstrapped estimates $\log \hat{p}^{*(b)}({X}=\tilde{\mathbf{x}})$ and $H_{Y|X}^{*(b)}(\tilde{\mathbf{x}})$ is actually a ways to assess the exogeneity condition.
 % \textcolor{red}{[Algorithm~\ref{} goes here...]}
Algorithm~\ref{alg:algoirithm} sunmmarizes the proposed procedure to determine the causal direction between $X$ and $Y$, given the sample $(\mathbf{x}, \mathbf{y})$ as input. In particular, it involves the following two modules.
% Algorithm 2!!!
\begin{small}
\begin{algorithm}[tb]
   \caption{Finding causal direction between $X$ and $Y$ based on exogeneity}
   \label{alg:algoirithm}
\begin{algorithmic}[0]
   \State {\bfseries Input:} data $(\mathbf{x}, \mathbf{y})$%, number of bootstrap replications $B$
   \State {\bfseries Output:} three possibilities: causal direction between $X$ and $Y$, or non-identifiable causal direction by exogeneity, or existence of hidden confunders
% \STATE consider hypothesis $X\rightarrow Y$
\State \Call{If$\_$Exogeneity}{$X \rightarrow Y$}
\State \Call{If$\_$Exogeneity}{$Y \rightarrow X$}
\If{exogeneity holds for only one direction}
\State {\bf return} the direction in which exogeneity holds
\ElsIf{exogeneity holds for both directions}
\State {\bf print} non-identifiable causal direction by exogeneity
\Else\Comment{exogeneity does not hold in either direction}
\State {\bf print} confounder case
\EndIf
\Statex
\Procedure{If$\_$Exogeneity}{$X\rightarrow Y$}
\For{$b = 1$ to $B$}
\State draw bootstrap resample $(\mathbf{x}^{*(b)}, \mathbf{y}^{*(b)})$ by random sampling with replacement from $(x_i, y_i)$;
\State estimate $\hat{p}^{*(b)}_X(X=\tilde{\mathbf{x}})$ and $H^{*(b)}_{Y|X}(\tilde{\mathbf{x}})$ with methods given in Sec.~\ref{Sec:estimation}
\EndFor
\State test for independence between $\hat{p}^{*(b)}_X(X=\tilde{\mathbf{x}})$ and $H^{*(b)}_{Y|X}(\tilde{\mathbf{x}})$, $b=1,...,B$, with the method given in Sec.~\ref{Sec:test}
\State \textbf{return} independence test result
\EndProcedure
\end{algorithmic}
\end{algorithm}
\end{small}

\subsubsection{Module 1: Nonparametric Estimators of $p(X)$ and $p(Y|X)$} \label{Sec:estimation}
When testing for exogeneity, one assumes the (parametric) model is correctly specified.  Otherwise, if the model is over-simplified, the estimated conditional distribution will depend on the marginal, which inspires the importance-reweighting scheme to handle learning problems under covariate shift (see e.g., Footnote 1 in~\cite{Sugiyama08}). For example, let us consider the situation where $Y$ depends on $X$ in a nonlinear manner while a linear model is exploited to estimate $p_{Y|X}$; clearly the estimate of the parameters in the conditional model would depend on that in $p_{X}$. %\textcolor{red}{[depends on the estimate or the parameter itself?]}
To avoid this, we use flexible nonparametric models %, such as %the mixture of Gaussians to estimate the marginal distribution and
%mixture of Gaussian processes (GPs)
to estimate the conditional.
\begin{comment}In this way, we can not directly access the parameters $ \theta$ and $ \psi$; however, similarly as the parametric case, we can test for the independence between $\Delta \log p_X$ and $\Delta  \mathbb{E}_{Y|X} \log p_{Y|X}$ instead, as shown below.
\end{comment}

% {\bf Nonparametric estimators of $p(X)$ and $p(Y|X)$. }
Suppose we aim to verify if $X$ exogenous for effective ``parameters" in $P(Y|X)$.
We need to estimate the marginal distribution $p(X)$ and the conditional distribution $p({Y|X})$ on the original sample as well as each bootstrap resample. %There are a number of approaches to do so. %For instance, kernel-density estimation is widely used; considering the conditional density as the ratio of the joint distribution to the marginal one, one can adopt the method proposed in [Sugiyama?] for conditional density estimation.
%\subsection{With explicit density estimation}
We estimate $p(X)$ with Gaussian kernel density estimation, and the kernel width was selected by Silverman's rule of thumb~\cite[page 48]{Silverman98}.

To estimate the conditional density $p(Y|X)$, we adapted the method orignally proposed for causal inference based on the structural equation $Y=f(X,E)$~\cite{Mooij10_GPI}. This method aims to find the functional causal model $Y = f(X,E)$, where $E\independent X$, given $(\mathbf{x}, \mathbf{y})$.  Without loss of generality, one can assume that $E \sim \mathcal{N}(0,1)$. (Otherwise, one can always write $E=g(\tilde{E})$ where $g$ is some appropriate function and $\tilde{E} \sim \mathcal{N}(0,1)$, and use the functional causal model $Y=f\big(X,g(\tilde{E})\big)$ instead.)  Here $f$ is completely nonparametric: it takes a Gaussian process prior with zero mean function and covariance function $k\big((x,e), (x',e') \big)$, where $k$ is a Gaussian kernel, and $(x,e)$ and $(x',e')$ are two points of $(X,E)$. Like in~\cite{Lawrence05_GPLVM}, this method optimizes the values of $E$, denoted by $\hat{e}_i$, as well as involved hyperparameters, and produces the maximum a posterior (MAP) solution of $f$, by maximizing the approximate marginal likelihood. The functional causal model implies the conditional density:
$$P(Y|X) = \frac{p(X,Y)}{p(X)} = \frac{p(X,E)\big/|\frac{\partial f}{\partial E}|}{p(X)}  = p(E)\Big/|\frac{\partial f}{\partial E}|.$$
Finally, once we have the $\hat{e}_i$ and the estimate of $f$, the conditional density at each point can be estimated as $p(Y =y_i | X = x_i) = p(E = \hat{e}_i)/\Big| \frac{\partial f}{\partial E}(x_i, \hat{e}_i)\Big|$.

%On each bootstrap resample, we use that method to estimate $\Big|\frac{\partial f}{\partial E} \Big|$ and estimate $\hat{e}_i$, and then the conditional distribution. %can be estimated as $p({Y|X}) = p(E)/|\frac{\partial f}{\partial E}|$, where $P(E)$ is specified in advance, and we use the standard Gaussian distribution for it.

\begin{comment}
In fact, one can think of the functional causal model as a way to specify the conditional distribution.  Correspondence... (One can always find the functional causal model...)  Jacobian... An efficient way to represent the conditional density...

\begin{equation} \label{Eq:conditional_density} p_{Y|X} = p_E/|\frac{\partial f}{\partial E}|,
\end{equation}
where $p_E$ is specified in advance; as in [], we use the standard Gaussian distribution for it.
\end{comment}

%\subsection{Testing for independence}% between $\Delta \log \hat{p}_X$ and $\Delta  \mathbb{E}_{Y|X} \log \hat{p}_{Y|X}$ on bootstrap samples}

% {\bf Testing for independence. }
\subsubsection{Module 2: Testing for Independence Between High-Dimensional Vectors} \label{Sec:test}

The task is then to test for independence between the estimated quantities on the bootstrap resamples, $\log \hat{p}^{*(b)}(X=\tilde{\mathbf{x}})$ and $H_{Y|X}^{*(b)}(\tilde{\mathbf{x}})$, $b=1,...,B$. Their dimentions are the number of data points in $\tilde{\mathbf{x}}$, which is 80 in our experiments.

%Suppose we generate $B$ bootstrap samples.
Let $\mathbf{R}$ be the matrix consisting of the centered version of $\log \hat{p}^{*(b)}(X=\tilde{x}_i)$, obtained on all bootstrap resamples, i.e., the $(i,b)$th entry of $\mathbf{R}$ is
$$R_{ib} \triangleq \log \hat{p}(X^{*(b)}(X=\tilde{x}_i) - \frac{1}{B}\sum_{k=1}^B \log \hat{p}^{*(k)}(X=\tilde{x}_i).$$  Similarly, $\mathbf{S}$ contains the centered version of $H_{Y|X}^{*(b)}(\tilde{x}_i)$, i.e.,
$$S_{ib} \triangleq H_{Y|X}^{*(b)}(\tilde{x}_i) - \frac{1}{B}\sum_{k=1}^B H_{Y|X}^{*(k)}(\tilde{x}_i).$$

%{\bf Permutation test}
Both $\mathbf{R}$ and $\mathbf{S}$ are of the size $N\times B$. We define the statistic as $C_{X\rightarrow Y} \triangleq \textrm{Tr}\big((\mathbf{RS}^T)(\mathbf{RS}^T)^T\big) = \textrm{Tr}(\mathbf{R}^T\mathbf{R}\cdot \mathbf{S}^T\mathbf{S})$, which is actually the sum of squares of the covariances between all rows of $\mathbf{R}$ and those of $\mathbf{S}$. The distribution of this statistic under the null hypothesis that $\log \hat{p}^{*(b)}(X=\tilde{\mathbf{x}})$ and $H_{Y|X}^{*(b)}(\tilde{\mathbf{x}})$ are independent can then be constructed by permutation test.

Note that this statistic is actually the Hilbert-Schmidt independence criterion (HSIC)~\cite{GreBouSmoSch05} with a linear kernel. That is, we care about linear dependence between $\log \hat{p}^{*(b)}(X=\tilde{\mathbf{x}})$ and $H_{Y|X}^{*(b)}(\tilde{\mathbf{x}})$; this is reasonable because they are in the vicinity of the maximum likelihood estimates and their dependence can be captured by linear approximation. On the other hand, if we use HSIC with Gaussian kernels, the result will be sensitive to the kernel width because the data dimension (the number of rows of $\mathbf{R}$ and $\mathbf{S}$) is high.

\section{Experiments}

In this section we first evaluate the behavior of the proposed bootstrap-based method for causal inference with synthetic data, on which the ground-truth is known, and then apply it on real data.  We use two variables, and with synthetic data, we examine both the case where the two variables have a direct causal relation and the confounder case (i.e., there are confounders influencing both of them). We compare the proposed bootstrap-based approach with the additive noise model (ANM) proposed in~\cite{Hoyer09}), GPI~\cite{Mooij10_GPI}, and information-geometric causal inference (IGCI) approach~\cite{Janzing12}: ANM assumes that the effect is a nonlinear function of the cause plus additive noise, GPI applies the Gausian Process prior on the generating function, and IGCI assumes the transformation from the cause to the effect is deterministic, nonlinear, and independent from the distribution of the cause in a certain way.  For computational reasons, we used 1000 bootstrap replications.%%%%
%\begin{small}
\begin{figure}[htp] 
\begin{center}
%\begin{minipage}{.31\linewidth}
\includegraphics[width = .74\linewidth]{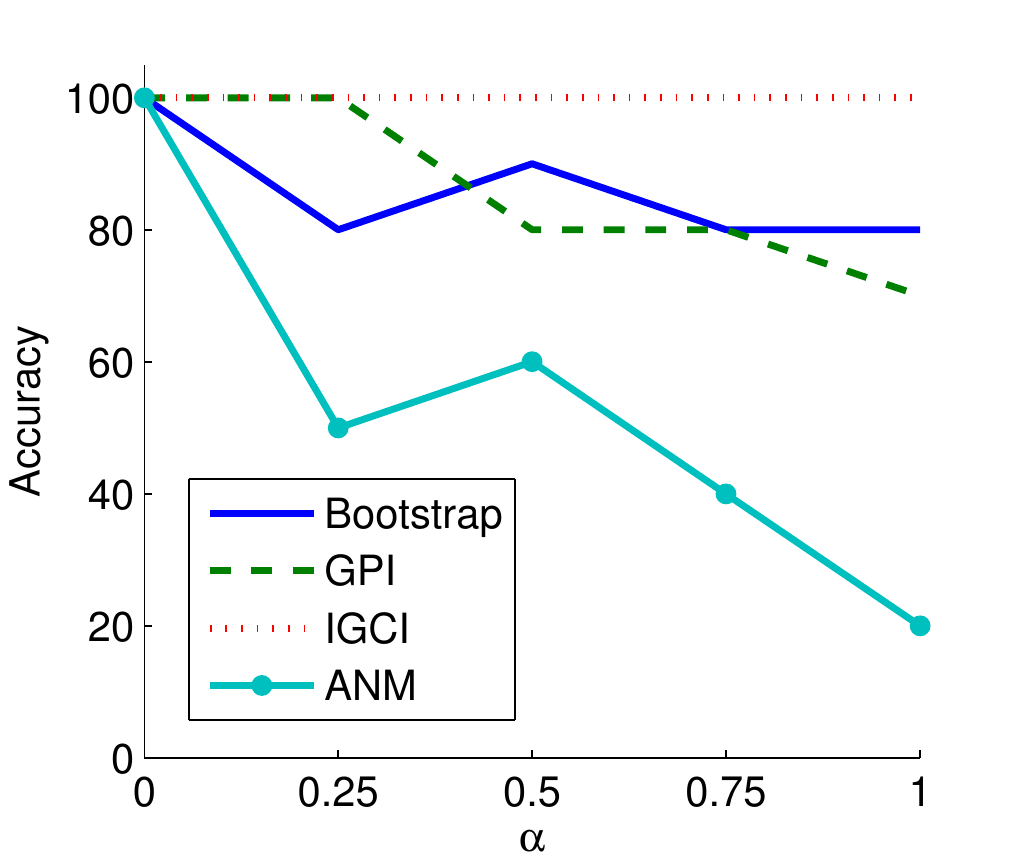} 
\\(a) Changing $\alpha$: From additive to multiplicative noise %\end{minipage} ~~~
%\begin{minipage}{.31\linewidth}~~~~
\\ \includegraphics[width = .74\linewidth]{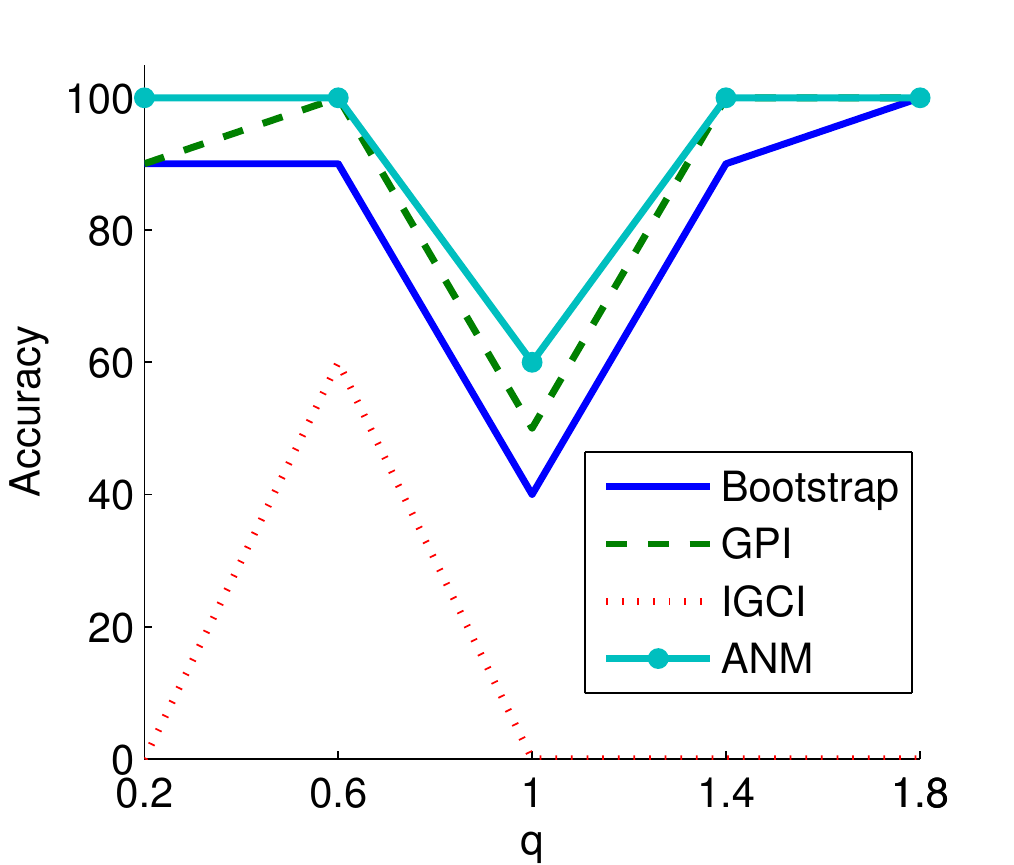} \\ (b) Changing $q$: From sub-Gaussian to super-Gaussian additive noise %\end{minipage} ~~~%\\
%\centering\begin{minipage}{.31\linewidth}\centering
\\ \includegraphics[width = .74\linewidth]{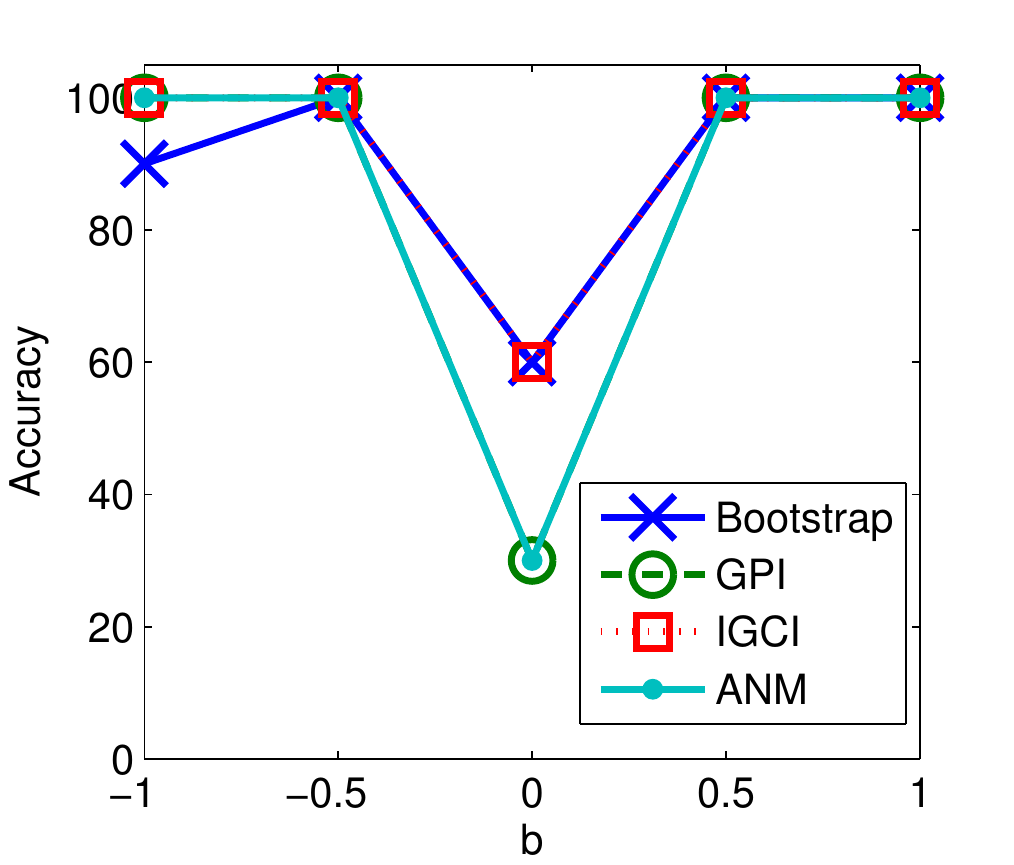}\\ (c) Changing $b$: Various nonlinear functions with Gaussian additive noise
%\end{minipage}
\end{center} \vspace{-0.2cm}
\caption{Accuracy of correctly estimating the causal direction for different generating models: (a) $q=1$, $b=1$, and $\alpha$ changed from 0 to 1, (b) for a linear function ($b=0$) with additive noise ($\alpha=0$) which changed from sub-Gaussian ($q<1$) to sub-Gaussian ($q>1$), and (c) various nonlinear functions ($b$ changed from -1 to 1) with additive Gaussian noise ($q=1$, $\alpha=0$).}
\label{fig:result_simulation1}\end{figure}% \end{small}
\vspace{-0.2cm}

%\subsection{Simulation: Without confounders}
{\bf Simulation: Without Confounders. }
Inspired by the settings in~\cite{Hoyer09,Mooij10_GPI}, we generated the simulated data with the model $Y = (X+bX^3)e^{\alpha E} + (1-\alpha)E$, where $X$ and $E$ were obtained by passing i.i.d. Gaussian samples through power nonlinearities with exponent $q$, while keeping the original signs. The parameter $\alpha$ controls the type of the observation noise, ranging from purely additive noise ($\alpha=0$) to purely multiplicative noise ($\alpha = 1$). $b$ determines how nonlinear the effect of $X$ is, and when $b=0$ the model is linear. The parameter $q$ controls the non-Gaussninity of $X$ and $E$: $q=1$ corresponds to a Gaussian distribution, and $q>1$ and $q<1$ produce super-Gaussian and sub-Gaussian distributions, respectively.

We considered three situations, in each of which two of $q$, $b$, and $\alpha$ were fixed and we see how the other changes the performance of different methods. For each combination of $q$, $b$, and $\alpha$, we independently simulated 10 data sets with 500 data points.\footnote{Since the bootstap-based approach is rather time-consuming, we only simulated 10 data sets for each setting.}  Fig.~\ref{fig:result_simulation1} shows the accuracy of the considered methods. One can see that the accuracy of the bootstrap-based approach is among or close to the best results, %does not necessarily perform best,
indicating that it is able to perform causal inference in various situations. We note that in practice, the performance of the bootstrap-based approach depends on the number of bootstrap replications and the method used for conditional distribution estimation. Although due to computatioanl reasons, we did not try a larger number of bootstrap replications, generally speaking, the accuracy of the bootstrap-based method improves as the number of replications increases.

\begin{figure}[ht]
%\begin{center}
%\begin{small}
\begin{center}%\begin{minipage}{.45\linewidth}\centering
\includegraphics[width = 1\linewidth]{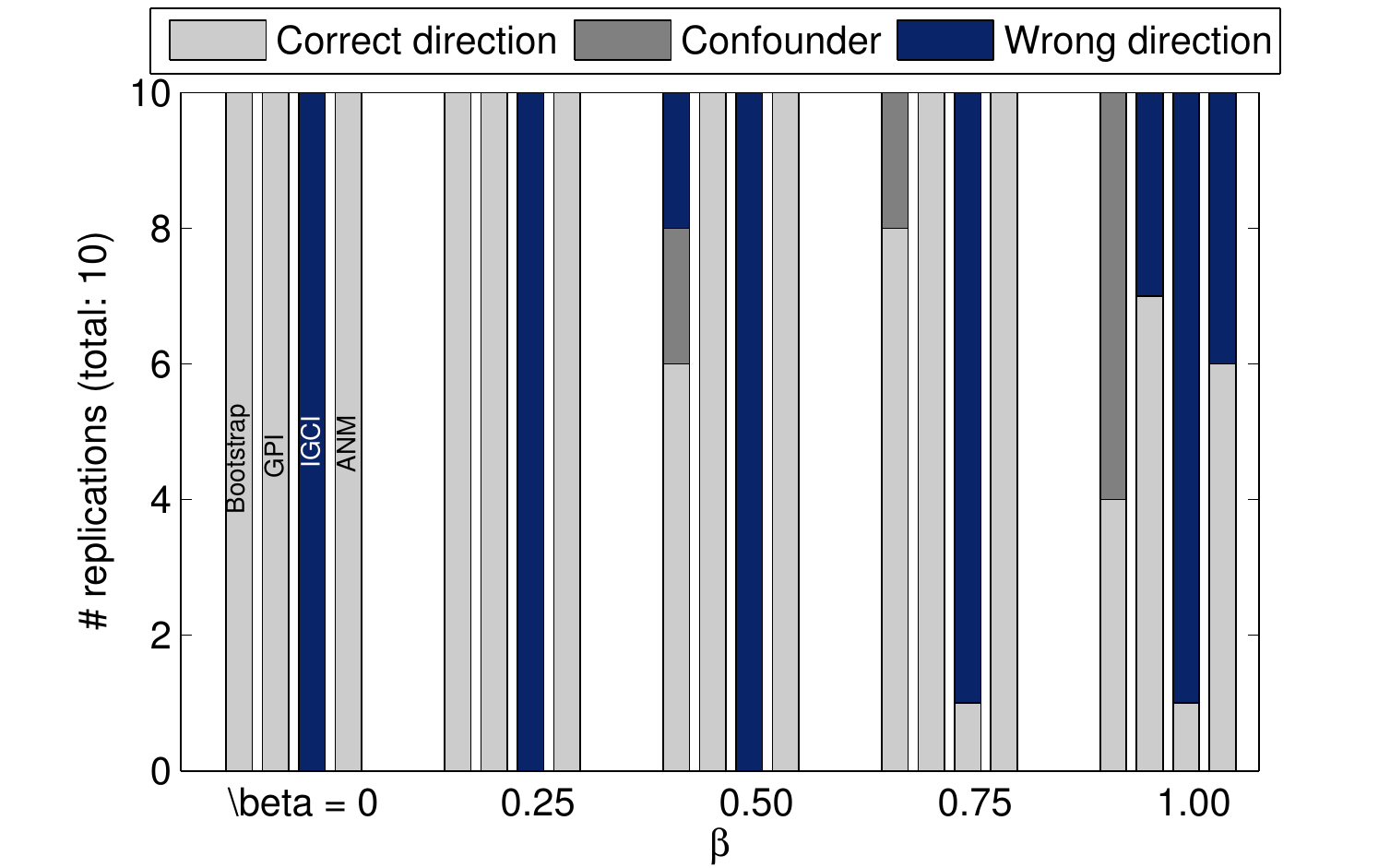}\\(a) Situation 1: Linear confounder case.\\~ %\end{minipage} %\\
%\vspace{.2cm}
%\begin{minipage}{.45\linewidth} \centering 
\\ \includegraphics[width = 1\linewidth]{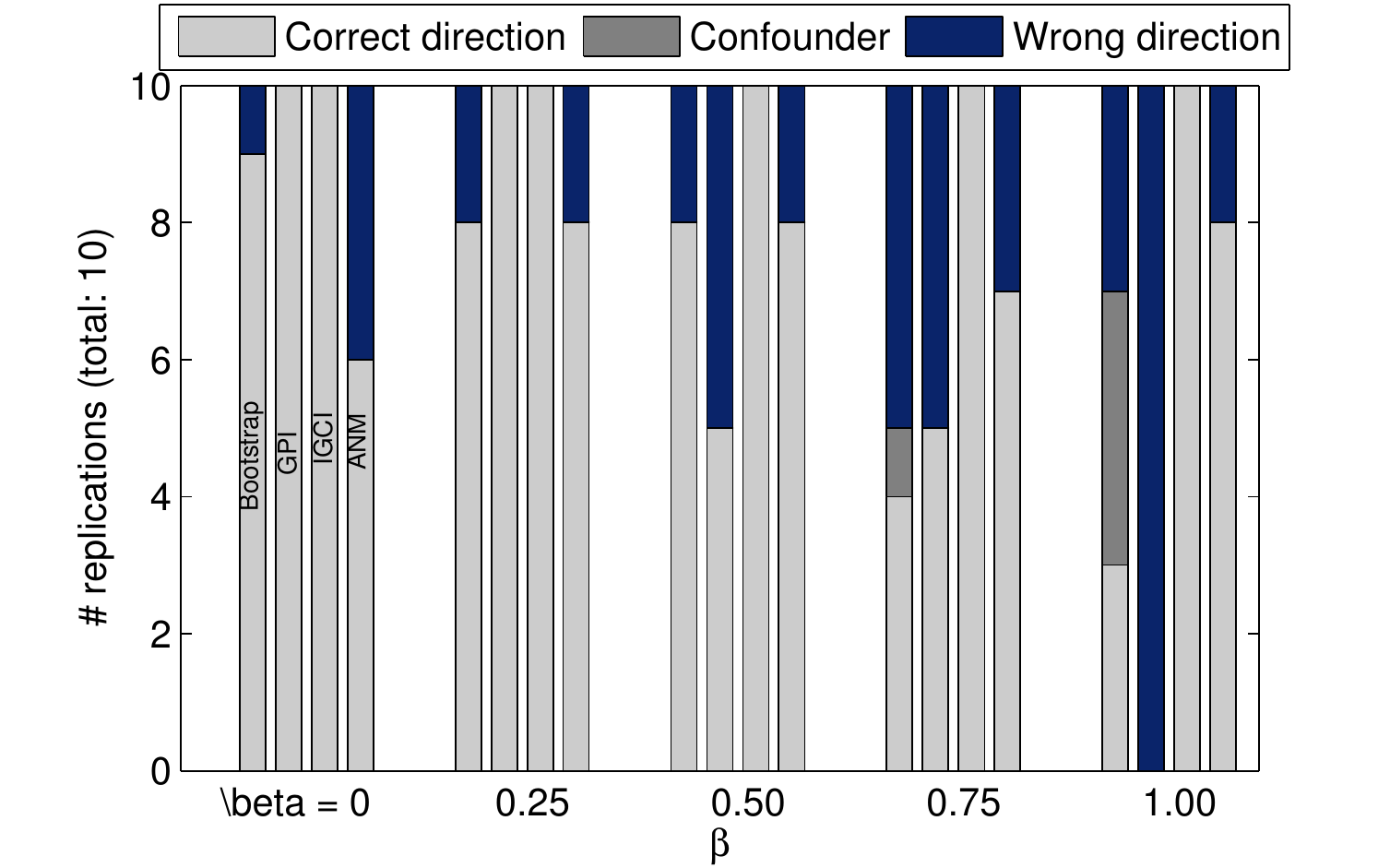} \\
(b) Situation 2: Nonlinear confounder case.
% \end{minipage}
\end{center} \vspace{-0.2cm}
\caption{Number of replications in which the methods find correct directions, report existence of confounders, and give wrong directions, respectively. {\it For each value of $\beta$, the four bars correspond to bootstrap-based method, GPI, IGCI, and ANM (from left to right).}}
\label{fig:sim_confounder}
\end{figure}

%\subsection{Simulation: With confounders}
{\bf Simulation: With Confounders. }
We then include the confounder variable $Z$ in the system, so that the causal structure is $Z\rightarrow X$ and $(Z,X) \rightarrow Y$.  For simplicity, we assume that both $X$ and $Y$ are influenced by $Z$ in a linear form: $X = (2-\beta) E_X + \beta Z$, and $Y = 0.3(2-\beta)\big[(X+bX^3)e^{\alpha E} + (1-\alpha)E \big] + \beta Z$, where $E_X$, $Z$, and $E$ were obtained by passing i.i.d. Gaussian samples through power nonlinearities with exponent $q = 1.5$, and $\beta$ controls how strong the effect of $Z$ is on both $X$ and $Y$.  We considered two situations: in one of them, we set $\alpha = 0$ and $b = 0$, i.e., the whole model is linear; in the other situation, $\alpha = 0.2$, and $b= 0.3$, so the model contains both additive noise and multiplicative noise. We changed $\beta$ from 0 to 1, and Fig.~\ref{fig:sim_confounder} shows the performances of the four methods in the two situations; note that for each value of $\beta$, the four bars (from left to right) correspond to the bootstrap-based method, GPI, IGCI, and ANM.  In particular, one can see that the bootstrap-based method  tends to detect the presence of the confounder when its effect is significant.

%  We assume $Z$ is standard Gaussian, and $X$ and $Y$ are generated as follows...... where $\gamma$ controls how significant the contribution of $Z$ is on both $X$ and $Y$......

% \begin{small}
% \end{small}
% \vspace{-0.2cm}

%\subsection{On real data}
{\bf On Real Data. }
We applied the bootstrap-based method on the cause-effect pairs available at  \\ \url{http://webdav.tuebingen.mpg.de/cause-effect/}. \\To reduce computational load, we used at most 500 points for each cause-effect pair. On 20 pairs (pairs 21, 43, 45, 48-51, 56-58, 61-64, 72, 75, 77-79, and 81), the p-values of the independence test for both directions are smaller than 0.01, indicating that there might be significant confounders. This seems reasonable, as the data scatter plots for these pairs indicate that the two variables have complex dependence relationships.
On the remaining 57 data sets, the bootstrap-based method output correct causal directions on 41 of them (with an accuracy 72\%). We also applied the recently proposed causal inference approaches, including IGCI~\cite{Janzing12}, the approach based on the Gaussian process prior~\cite{Mooij10_GPI}, and that based on the post-nonlinear causal model~\cite{Zhang_UAI09} on those 57 data sets for comparison. Their performance was similar: the three approaches gave correct causal directions on 41, 40, and 43 pairs, respectively.
%; among them, on 6 data sets (#) both directions seem possible as the p-values for both directions are bigger than 0.05

\begin{comment}
\subsubsection{Two Modules: Estimating the Marginal and Conditional Distributions}
Suppose we aim to see if $X$ causes $Y$.
We need to estimate the marginal distribution $p_X$ and the conditional distribution $p_{Y|X}$ on the original given sample as well as each bootstrap sample. For the former, we use the mixture of Gaussians. For the latter, we adopt the mixture of Gaussian processes (in particular, the overlapping mixtures of Gaussian Processes []). \textcolor{red}{No! It seems that this method works poorly when data are very noisy.}
\end{comment}

\section{Conclusion and discussions}

%If we estimate the conditional distribution $p_{Y|X}$ separately on each point of $X$, this method does not apply, because there are no shared parameters... each one is estimated independently...  always independent(???)...  Compact representation of the conditional distribution...

We proposed to do causal inference based on the criterion of exogeneity of the cause for the parameters in the conditional distribution of the effect given the cause. We discussed how to assess such exogeneity in nonparametric settings. To this end, one needs to draw a number of samples according to the unknown data-generating process. Fortunately, the bootstrap provides a way to mimic the data generating process from which we can draw a number of samples and analyze their statistical properties.

%We also found that the method works well even  if the conditional density cannot be estimated accurately...

Our approach shows that it is possible to determine causal direction without structural constraints or a specific type of smoothness assumptions on the functional models. The proposed computational approach successfully
 demonstrated the validity of this idea, though it is computationally demanding because of the bootstrap procedure and its performance is not necessarily the best among existing methods. At the same time, it enjoys some advantages. First, it does not make a strong assumption on the data-generating process. Second, it could often tell us if significant confounders exist. The performance of the proposed bootstrap-based approach depends on the number of bootstrap replications and the method for conditional distribution estimation. In future work we aim to develop more reliable methods along this line, including methods that can handle more than two variables.

\textcolor {black}{In this paper we made an attempt to discover causal information from observational data based on a condition of exogeneity, which provides another perspective to conceptualize the "independence" between the process generating the cause and that generating the effect from cause.  On the other hand, it is worth mentioning that this type of independence is able to facilitate understanding and solving some machine learning or data analysis problems. For instance, it helps understand when unlabeled data points will help in the semi-supervised learning scenario~\cite{Scholkopf12}, and inspired new settings and formulations for domain adaptation by characterizing what information to transfer and how to do so~\cite{Zhang13_targetshift,Zhang15_multi}.}

\section*{Acknowledgement}
We would like to thank Peter Spirtes, Kevin Kelly, and Dominik Janzing for helpful discussions.  K. Zhang was supported in part by DARPA grant No. W911NF-12-1-0034.  The research of J. Zhang was supported in part by the Research Grants Council of
Hong Kong under the General Research Fund LU342213.

%\begin{small}
% \bibliography{sigproc}
%\bibliographystyle{unsrt}
%% \bibliographystyle{abbrv}
%% \bibliography{term_fin1}
%\end{small}

 \newpage
\begin{comment}{
 \begin{onecolumn}
  \centering{\LARGE Supplement to \\``Domain Adaptation under Target and Conditional Shift: A Kernel Mean Matching Approach"}

 \vspace{.6cm} {\large This supplementary material provides the
 proofs and some details which are omitted in the submitted paper. The equation
 numbers in this material are consistent with those in the paper.}
 \vspace{.3cm}
 \end{onecolumn}
}
\end{comment}

\twocolumn[\begin{@twocolumnfalse}
\begin{center}{\LARGE Supplement to \\``Distinguishing Cause from Effect Based on Exogeneity"}\end{center} \vspace{1.5cm}

\end{@twocolumnfalse}]

 \vspace{.6cm} {\large This supplementary material provides the
 proofs and discussions which are omitted in the submitted paper. The equation
 numbers in this material are consistent with those in the paper.}\\~~\\

\section*{S1. Mutual Exogeneity and Its Relationship to Definition~\ref{Defi:exo_cut}}
% mutual exogeneity. relationship to Bayesian cut
There are two types of analysis of exogeneity~\cite{Florens85}; one considers the inference based on the complete sample results, and the other considers dynamic models where the data were obtained by ``sequential sampling". In this paper we focus on the former scenario.

From the Bayesian point of view, exogeneity of $X$ for $\psi$ allows an admissible reduction of the complete model $p(X,Y|\theta, \psi)$ to the conditional model $p(Y|X, \psi)$, in that both models lead to he same posterior distribution on the parameter set $\psi$~\cite{Florens85,Mouchart04}.  Below we give the definition of mutual exogneneity according to~\cite{Florens85}.

\begin{Defi}[Mutual exogeneity] \label{Def:mutual_exo}
$X$ and $\psi$ are mutually exogenous if and only if
\begin{enumerate}
\item [(i)] $\psi$ and $X$ are independent, i.e., $\psi \independent X$, and
\item [(ii)] $\psi$ is sufficient in the conditional distribution of $Y$ given $X$, i.e., $\theta \independent Y|(\psi,X)$.
\end{enumerate}
\end{Defi}
Here condition {\it (i)} is to do with the independence between $\psi$ and $X$; those two quantities play different roles in the model $p(X,Y,\psi, \theta)$, and consequently this independence condition is usually not convenient to verify.  Moreover, for the same reason, there is no fully equivalent concept in sampling theory (it is weaker than exogeneity defined in Definition~\ref{Defi:exo_cut}, because the property of $\theta$ is not specified).  A natural way of obtaining the mutual exogeneity of $X$ and $\psi$ is to exploit a stronger but more operational condition, namely the condition of the Bayesian cut.

\begin{comment}
% Definition of Bayesian cut []...
\begin{Defi}[Bayesian cut]\label{Def:cut}
$[\psi, (X, \theta)]$ operates a Bayesian cut in $p(X,Y,\theta,\psi)$ if
\begin{enumerate}
\item [(i)] $\psi$ and $\theta$ are independent {\it a priori}, i.e., $\psi \independent \theta$,
\item [(ii)] $\theta$ is sufficient in the sampling marginal process of $X$, i.e., $\psi\independent X|\theta$, and
\item [(iii)] $\psi$ is sufficient in the conditioal distribution of $Y$ given $X$, i.e., $\theta \independent Y|(\psi,X)$.
\end{enumerate}
\end{Defi}
\end{comment}

A Bayesian cut allows a complete separation of inference (on parameters $\theta$) in the marginal distribution and of inference (on $\psi$) in the conditional one. The prior independence between $\theta$ and $\psi$ in the Bayesian cut is a counterpart to the variation-free condition in the classical cut (condition {\it (ii)} in Definition~\ref{Defi:exo_cut}), and the last two conditions in Definition~\ref{Def:cut} implies condition {\it (i)} in Definition~\ref{Defi:exo_cut}. Thus, the Bayesian cut is equivalent to the classical cut in sampling theory, and consequently characterizes the exogeneity property defined in Definition~\ref{Defi:exo_cut}. Therefore, hereafter the exogeneity of $X$ for $\psi$ is used interchangeably with the statement that $[\psi, (X, \theta)]$ operates a Bayesian cut in $p(X,Y,\theta,\psi)$.  %We note that
%Note that in Economics sometimes the following condition in the above definition is replaced by tha condition that $\psi$ and $\theta$ are variation free.

The following theorem, extracted from~\cite{Florens90}, relates the Bayesian cut to the independence of the parameters according to the posterior distribution, as well as mutual exogeneity.
% also refer to book "Bayesian Inference in Dynamic Econometric Models"
\begin{theo1} \label{Theo:prior_posterior}
Suppose $[\psi, (X, \theta)]$ operates a Bayesian cut in $p(X,Y,\{\psi,\theta\})$; then
\begin{itemize}
\item [(i)] $X$ and $\psi$ are mutually exogenous, and
\item [(ii)] $\psi$ and $\theta$ are independent a posteriori.
\end{itemize}
On the other hand, if $X$ and $\psi$ are mutually exogenous and if $\theta \independent \psi |X$, $[\psi, (X, \theta)]$ operates a Bayesian cut.
\end{theo1}
When one (or more) condition in Definition~\ref{Def:cut} is violated, $[\psi, (X,\theta)]$ does not operate a Baysian cut, i.e., $X$ is not exogenous for $\psi$. Fig.~\ref{fig:data_generating}(b--d) shows the situations where conditions {\it(i)}, {\it(ii)}, and {\it(iii)} are violated, respectively, so that $[\psi, (X,\theta)]$ does not operate a Baysian cut. Note that by reparameterization, the three situations can reduce to each other. Take situations (b) and (c) as an example. If we divide $\theta$ in (b) into $(\theta_\gamma, \theta_{\independent})$, where $\theta_\gamma$ depends on $\gamma$ while $\theta_{\independent}$ does not, and consider $\theta_{\independent}$ as the new $\theta$, (b) becomes (c). Similarly, if we merge $\gamma$ and $\theta$ in (c) as the new $\theta$, we then have (b).
In all those situations, $\theta$ and $\psi$ are not independent {\it a posterior}, or the maximum likelihood estimates $\hat{\theta}$ and $\hat{\psi}$ are not independent according to the sampling distribution.

% \section*{S2. More Detail on ``An Identifiable Situation: Linear Non-Gaussian Case"}
\section*{S2. Relation to SEM-Based Causal Inference}

\subsection*{S2.1. Relation to Causal Inference Based on Marginal Likelihood}

%Functional causal models...

Recently, SEM-based approaches have demonstrated their power for causal inference of real-world problems. Structural equations represent the effect as a function of the causes and independent noise, which, from another point of view, provide a way to represent the conditional distribution $P(\texttt{effect} | \texttt{cause})$, or the causal mechanism. The generation of the cause-effect pair consists of two stages, one generating the cause according to $P(\texttt{cause})$ and the other further generating the effect from the value of the cause according to the structural equation.  The ``simplicity" constraints (e.g., linearity in~\cite{Shimizu06}, additive noise in~\cite{Hoyer09}, the post-nonlinear process in~\cite{Zhang_UAI09}, and the smoothness assumption in~\cite{Mooij10_GPI}) on the functions are crucial. On the one hand, they make the models asymmetric in cause and effect; otherwise, for any two variables, we can always represent one of the variables as a function of the other and an independent noise term~\cite{Hyvarinen99}. On the other hand, if the functions are constrained to be simple, the independence between the cause and the error terms would imply the exogeneity of the cause for the parameters in $P(\texttt{cause})$, as suggested by the error-based definition of exogeneity~\cite{Orcutt52} (see also~\cite{Pearl00}).\footnote{An error-based definition of exogeneity was given by~\cite{Orcutt52} (see also~\cite{Pearl00}): $X$ is said to be exogenous for parameters in $p(Y|X)$ is $X$ is independent of all errors that influence $Y$, except those mediated by $X$. We know that without appropriate constraints on the functions, given any two random variable, we can always represent one of them as a function of the other variable and an independent noise term~\cite{Hyvarinen99}, i.e., the functional causal models are not identifiable. Therefore, generally speaking, the above error-based definition is consistent with Definition~\ref{Defi:exo_cut} only when the functional class is well constrained. % We can see that if the functional form is unrelated to the marginal distribution of the assumed cause, the above definition is consistent with the one of exogeneity in~\cite{Engle83_exogeneity}.  However,
Otherwise, if the function and the distribution of the assumed cause are related in some way, the above definition is not rigorous. %\textcolor{red}{[Show this!]}
}

% and consequently physically interpretable, one can expect that the parameters involved in the two stages would be unrelated.
%****Cite the theorem here***

\begin{comment}
We note that this is implied in some concepts in philosophy and economics, such as ``invariance"~\cite{Woodward03} of the conditional distribution $P(\texttt{effect} | \texttt{cause})$ w.r.t. change in the marginal distribution $P(\texttt{cause})$ and ``exogeneity"~\cite{Engle83_exogeneity} of the cause for parameters in the conditional distribution $P(\texttt{effect} | \texttt{cause})$. Functional causal models, together with the simplicity assumption of the functions, offer a way to enforce such knowledge. In addition to providing a means of understanding why those methods perform so well on causal direction determination, those concepts can further inspire causal inference methods that would work in more generic situations.
\end{comment}

%More directly,
The concept ``exogeneity" provides theoretical support for the SEM-based causal inference methods that find the causal direction by comparing the marginal likelihood of the models in two directions; for an example of such methods, see~\cite{Mooij10_GPI}.\footnote{Note that due to computational difficulties, this method doe snot evaluate the marginal likelihood, but approximate it wiht the maximum regularized likelihood.} %In fact,  % allows us to determine the causal direction by comparing the marginal likelihoods of two different models.
One candidate model is given in Fig.~\ref{fig:data_generating}(a), where $X$ is exogenous for $\psi$ (or $[\psi, (X,\psi)]$) operates a Bayesian cut in $p(X,Y,\theta,\psi)$, denoted by $\mathcal{M}_1$. The other corresponds to the factorization:
\begin{equation}\label{Eq:factor2}
 p(X,Y|\tilde{\theta}, \tilde{\psi}) = p(Y|\tilde{\theta}) p(X|Y, \tilde{\psi}),
 \end{equation} 
 where $[\tilde{\psi}, (Y,\tilde{\theta})]$ operates a Bayesian cut in $p(Y,X,\tilde{\theta},\tilde{\psi})$, denoted by $\mathcal{M}_2$. Note that under the above models, the marginal likelihood of $(X,Y)$ is the product of that of the conditioning variable and that of the conditional distribution. Ideally, if all the involved distributions are correctly specified, one would prefer the causal direction $X\rightarrow Y$ (resp. $Y\rightarrow X$) if $\mathcal{M}_1$ (resp. $\mathcal{M}_2$) gives a higher marginal likelihood.

%[A theorem to support causal inference based on marginal likelihood with exogeneity...]

\begin{theo1} \label{Theo:marginal}
Suppose that the two random variables $X$ and $Y$ are generated according to $\mathcal{M}_1$, and that the exogeneity-based causal model is identifiable. Let the prior distributions of the parameters be $p^*(\psi|\mathcal{M}_{1})$ and $p^*(\theta|\mathcal{M}_{1})$.   For the given sample $(\mathbf{X}, \mathbf{Y})$, let $p(\mathbf{X}, \mathbf{Y}|\mathcal{M}_{1})$ be the marginal likelihood, i.e.,
\begin{small}
\begin{flalign} \nonumber
&p(\mathbf{X}, \mathbf{Y}|\mathcal{M}_{1}) \\ \nonumber
 = &\prod_{i=1}^N \iint\! p( {X_i,Y_i} | \{\psi,\theta \} ) p^*(\theta|\mathcal{M}_{1}) p^*(\psi|\mathcal{M}_{1}) \ddd\theta \ddd\psi \\ \nonumber
= &\prod_{i=1}^N \int \! p( {X_i} |\theta ) p^*(\theta|\mathcal{M}_{1})\ddd \theta \cdot \prod_{i=1}^N\int\! p( {Y_i} |{X_i}, \psi )  p^*(\psi|\mathcal{M}_{1}) \ddd\psi \\ \nonumber
= &\prod_{i=1}^N p( {X_i} |\mathcal{M}_{1}) \cdot \prod_{i=1}^N p( {Y_i} |{X_i}, \mathcal{M}_{1}).\end{flalign}  \end{small}
Assume that by a one-to-one reparametrization we can represent $p(X,Y|\{\psi, \theta\})$ as $p(Y|\tilde{\theta})p(X|Y,\tilde{\psi})$, where $Y$ is not exogenous for $\tilde{\psi}$.  Let $p(\mathbf{X}, \mathbf{Y}|\mathcal{M}_{2})$ be the marginal likelihood of $\mathcal{M}_{2}$, i.e.,
\begin{small}\begin{flalign} \nonumber
&p(\mathbf{X}, \mathbf{Y}|\mathcal{M}_{2}) \\ \nonumber
=& \prod_{i=1}^N \iint\! p( {X_i,Y_i} | \{\tilde{\psi},\tilde{\theta} \} ) p^0(\tilde{\theta}|\mathcal{M}_{2}) p^0(\tilde{\psi}|\mathcal{M}_{2}) \ddd\tilde{\theta} \ddd\tilde{\psi}
%\\ \nonumber
%&= \prod_{i=1}^N \int\! p( {Y_i} | \tilde{\theta} ) p^0(\tilde{\theta}|\mathcal{M}_{2}) \ddd\tilde{\theta} \cdot \prod_{i=1}^N\int\! p({X_i}|{Y_i}, \tilde{\psi}) p^0(\tilde{\psi}|\mathcal{M}_{2}) \ddd\tilde{\psi}
\\ \nonumber =& \prod_{i=1}^N  p( {Y_i} | \mathcal{M}_{2}) \cdot \prod_{i=1}^N p({X_i}|{Y_i}, \mathcal{M}_{2}),
\end{flalign} \end{small}
 where $\tilde{\theta}$ and $\tilde{\psi}$ have independent priors. As the sample size $N$ goes to infinity, for any choice of  $p^0(\tilde{\theta}|\mathcal{M}_{2}) $ and $p^0(\tilde{\psi}|\mathcal{M}_{2}) $, $ p(\mathbf{X}, \mathbf{Y}|\mathcal{M}_1)$ is always greater than $p(\mathbf{X}, \mathbf{Y}|\mathcal{M}_2)$.
\end{theo1}
%% conditional on "reparametrization"
%% should I write down the difference? refer to the paper "mutual information......"
\begin{proof}
As the data were generated according to model $\mathcal{M}_1$, we have 
$$\mathbb{E}\log p(X,Y|\mathcal{M}_1) = \int p(X,Y|\mathcal{M}_1) \log p(X,Y|\mathcal{M}_1) dxdy.$$ Furthermore,
\begin{small}\begin{flalign} \nonumber
&\mathbb{E}\log p(X,Y|\mathcal{M}_1)  - \mathbb{E}\log p(X,Y|\mathcal{M}_2) \\ \nonumber
=& \int p(X,Y|\mathcal{M}_1) \log\frac{ p(X,Y|\mathcal{M}_1)}{p(X,Y|\mathcal{M}_2)} dxdy
\\ \nonumber =& \mathcal{D}\big(p(X,Y|\mathcal{M}_1)~||~p(X,Y|\mathcal{M}_2)\big),
\end{flalign} \end{small}
where $\mathcal{D}(\cdot || \cdot)$ denotes the Kullback-Leibler divergence.  Clearly the above quantity is non-negative, and it is zero if and only if $p(X,Y|\mathcal{M}_1) = p(X,Y|\mathcal{M}_2)$ for all possible $x$ and $y$. However, this condition cannot hold, because the model $\mathcal{M}_1$ is assumed to be identifiable based on exogeneity.

Consequently, we have $\mathbb{E}\log p(X,Y|\mathcal{M}_1)  > \mathbb{E}\log p(X,Y|\mathcal{M}_2)$. Moreover, according to the weak law of large numbers, as $N \rightarrow \infty$,   $\frac{1}{N} \log p(\mathbf{X}, \mathbf{Y}|\mathcal{M}_{1})$  and $\frac{1}{N} \log p(\mathbf{X}, \mathbf{Y}|\mathcal{M}_{2})$ will convergence in probability to the quantities $\mathbb{E}\log p(X,Y|\mathcal{M}_1)$ and $\mathbb{E}\log p(X,Y|\mathcal{M}_2)$, respectively. That is, if $N$ is large enough, $p(\mathbf{X}, \mathbf{Y}|\mathcal{M}_{1}) > p(\mathbf{X}, \mathbf{Y}|\mathcal{M}_{2})$.
\end{proof}
%\end{comment}
\begin{comment}XXXX
As reparameterization of the parameters does not change the likelihood, for correspondent values of $({\theta},{\psi})$ and $(\tilde{\theta},\tilde{\psi})$  we have $p(X,Y|\tilde{\theta},\tilde{\psi}) = p(X,Y|\tilde{\theta},\tilde{\psi})$. Consequently we have
\begin{eqnarray} \nonumber
p({X}, {Y}|\mathcal{M}_{1}) &=&  \int p({X,Y} | \{\psi,\theta \} ) p^*(\theta|\mathcal{M}_{1}) p^*(\psi|\mathcal{M}_{1}) d\theta d\psi \\ \label{eq_med1}
&=&  \int p( {X,Y} | \{\tilde{\psi},\tilde{\theta} \} ) p^*(\tilde{\theta},\tilde{\psi}) d\tilde{\theta} d\tilde{\psi},
\end{eqnarray}
where $p^*(\tilde{\theta},\tilde{\psi}) $ are the prior distribution of $(\tilde{\theta},\tilde{\psi})$ corresponding to the prior $p^*(\theta|\mathcal{M}_1)p^*(\psi|\mathcal{M}_1)$.

We then show that the quantity in~(\ref{eq_med1}) is not equal to $p(X,Y|\mathcal{M}_2)$.
\end{comment}

%\begin{comment}
However, the marginal likelihood depends heavily on the models or assumptions for the marginal and conditional distributions. Besides the exogeneity property, such approaches also make additional assumptions about the functions, such as structural constraints~\cite{Shimizu06,Hoyer09,Zhang_UAI09} and the smoothness assumption~\cite{Mooij10_GPI}.  The proposed approach avoids such assumptions, by directly assessing the exogeneity property.

%Only optimal when they do not share parameters and the posterior distribution has no coupling between learned parameters of the conditional and the input density functions... [Lars Kai hansen, page 68]

\subsubsection*{S2.1.1. A Simple Illustration on Parametric Models with Laplace Approximation}

%{\bf An illustration on parametric models with Laplace approximation. }
Here we use a somehow oversimplified parametric example to illustrate why the marginal likelihood implies the causal direction. Assume that $\mathcal{M}_1$ holds, that is, in factorization (\ref{Eq:factorization}), $X$ is exogenous to $\psi$. We will demonstrate that the likelihood for model (\ref{Eq:factor2}) would be asymptotically smaller if we wrongly assume that $Y$ is exogenous for $\tilde{\psi}$. We assume that there is a one-to-one correspondence between $(\theta, \psi)$ and $(\tilde{\theta}, \tilde{\psi})$. As seen from the proof of Theorem~\ref{Theo:marginal}, the marginal distribution of (\ref{Eq:factorization}) under $\mathcal{M}_1$ would be the same as that of (\ref{Eq:factor2}) with the dependence between $\tilde{\theta}$ and $\tilde{\psi}$ taken into account. Suppose that the corresponding log marginal likelihood  $\log p(\mathbf{X},\mathbf{Y}|\mathcal{M}_1)$, can be evaluated with the Laplace approximation in terms of $(\tilde{\theta}, \tilde{\psi})$~\cite{Kass88}:
\begin{comment}\begin{small}\begin{flalign} \nonumber
\log p(\mathbf{X},\mathbf{Y}|\mathcal{M}_1)
\approx & \log p(\mathbf{X},\mathbf{Y}|\hat{\theta}) + \log p^0(\hat{\theta}) + \log p^0(\hat{\psi}) \\ \nonumber &~~~~~~~~~~~~~~~~- \frac{1}{2}\log |\Sigma_{\theta,\psi}| + \frac{d}{2}\log(2\pi),
\end{flalign} \end{small}
\end{comment}
\begin{flalign} \nonumber
\log p(\mathbf{X},\mathbf{Y}|\mathcal{M}_1)
& \approx  \log p(\mathbf{X},\mathbf{Y}|\hat{\tilde{\theta}},\hat{\tilde{\psi}}) + \log p^0(\hat{\tilde{\theta}},\hat{\tilde{\psi}}) \\ \nonumber
&~~~~~~~~~~- \frac{1}{2}\log |\Sigma_{\tilde{\theta},\tilde{\psi}}| + \frac{d}{2}\log(2\pi),
\end{flalign}
where $\hat{\tilde{\theta}}$ and $\hat{\tilde{\psi}}$ are the maximum a posterior (MAP) estimate, $p^0(\hat{\tilde{\theta}},\hat{\tilde{\psi}})$ is the prior, $\Sigma_{\tilde{\theta},\tilde{\psi}}$ is the negative Hessian of $\log[p(\mathbf{X},\mathbf{Y}|\tilde{\theta},\tilde{\psi})p^0(\tilde{\theta})p^0(\tilde{\psi})]$ evaluated at $(\hat{\tilde{\theta}}, \tilde{\hat{\psi}})$, and $d$ is the number of parameters.

On the other hand, under $\mathcal{M}_2$, the negative Hessian matrix becomes $\tilde{\Sigma}_{\tilde{\theta},\tilde{\psi}}$  which is block-diagonal and shares the same main diagonal block matrices $\Sigma_{\tilde{\theta}}$ and $\Sigma_{\tilde{\psi}}$ with $\Sigma_{\tilde{\theta},\tilde{\psi}}$. We then have
$\log p(\mathbf{X},\mathbf{Y}|\mathcal{M}_1) - \log p(\mathbf{X},\mathbf{Y}|\mathcal{M}_2) \approx \frac{1}{2}\big(\log |\tilde{\Sigma}_{\tilde{\theta},\tilde{\psi}}| - \log |\Sigma_{\tilde{\theta},\tilde{\psi}}|\big) = \frac{1}{2}\big( \log |\Sigma_{\tilde{\theta}}| + \log |\Sigma_{\tilde{\psi}}| - \log |{\Sigma}_{\tilde{\theta},\tilde{\psi}}|\big).$
One can show that $|{\Sigma}_{\tilde{\theta},\tilde{\psi}}| < |\Sigma_{\tilde{\theta}}|\cdot |\Sigma_{\tilde{\psi}}|$ if $\Sigma_{\tilde{\theta},\tilde{\psi}}$ is not block-diagonal; for a proof, see \cite[page 239]{Numbertheory}. Hence, we have $\log p(\mathbf{X},\mathbf{Y}|\mathcal{M}_1) > \log p(\mathbf{X},\mathbf{Y}|\mathcal{M}_2)$ asymptotically.

%  the reparameterization from $(\theta, \psi)$ to $(\tilde{\theta}, \tilde{\psi})$ does not change the data likelihood As explained in section~\ref{Sec:exogeneity},

% A simple illustration with linear models?

\subsection*{S2.2. Relation to Invariance of SEMs}

The proposed bootstrap-based method provides a way to examine if an equation is structural or not.  Suppose $Y = f(X,E)$, where $E\independent X$, is a structural causal model in that $f$ is invariant to changes in the distribution of $X$~\cite{Pearl00}. One can then see that since $E$ and $X$ are independent processes, the bootstrapped $\hat{P}^{*(b)}(X)$ is independent from the underlying $\hat{p}^{*(b)}(E)$, and hence independent from $\hat{p}^{*(b)}(Y|X) = \hat{p}^{*(b)}(E)/\big|\frac{\partial f}{\partial E}\big|$.

Now consider the other direction. According to~\cite{Hyvarinen99}, we can always find an equation $X=\tilde{f}(Y;\tilde{E})$ such that $\tilde{E}\independent Y$; suppose this equation is not structural, in that $\tilde{f}$, or in particular, $\big|\frac{\partial \tilde{f}}{\partial \tilde{E}}\big|$ is dependent on $p(Y)$. Again, we have $\hat{p}^{*(b)}(X|Y) = \hat{p}^{*(b)}(\tilde{E})/\big|\frac{\partial \tilde{f}}{\partial \tilde{E}}\big|$. The bootstrapped $\hat{p}^{*(b)}(Y)$ and $\hat{p}^{*(b)}(X|Y)$ are then dependent due to the dependence between $\big|\frac{\partial \tilde{f}}{\partial \tilde{E}}\big|$ and $\hat{p}^{*(b)}(Y)$.

In particular, the SEM-based causal inference approaches~\cite{Shimizu06,Hoyer09,Zhang_UAI09,Mooij10_GPI} constrain the functions $f$ to be simple in respective senses; consequently they are not so flexible as to change with the input distribution $p(X)$, and then the independence between the input $X$ and the noise $E$ serves as a surrogate to achieve the exogeneity condition of $X$ for the parameters in $p(Y|X)$.

Compared to SEM-based approaches, the proposed exogeneity-based approach avoids the constraints on the functional causal model $f$.  On the other hand, some SEM-based approaches have clear identifiability conditions under which the reverse
direction $Y\rightarrow X$ that induces the same joint distribution on $(X,Y)$ does not exist in general, given the causal direction $X\rightarrow Y$; for instance, see~\cite{Hoyer09,Zhang_UAI09}. However, to find theoretical identifiability results for the proposed approach, one has to establish the identifiability conditions in terms of data distributions, which turns out to be extremely difficult.

\end{document}